\documentclass[%
preprint, showpacs,preprintnumbers,
 amsmath,amssymb,
 aps,
pre
]{revtex4-1}

\usepackage{graphicx}
\usepackage{dcolumn}
\usepackage{bm}
\usepackage[makeroom]{cancel}
\usepackage{natbib}

\begin{document}

\preprint{APS/123-QED}

\title{Information-theoretical analysis of the
statistical dependencies among three variables: Applications to
written language}

\author{Dami\'an G. Hern\'andez}
\author{Dami\'an H. Zanette}
\author{In\'es Samengo}

\affiliation{Centro At\'omico Bariloche and Instituto Balseiro,
(8400) San Carlos de Bariloche, Argentina.}


%

\date{\today}

\begin{abstract}
We develop the information-theoretical concepts required to study
the statistical dependencies among three variables. Some of such
dependencies are pure triple interactions, in the sense that they
cannot be explained in terms of a combination of pairwise
correlations. We derive bounds for triple dependencies, and
characterize the shape of the joint probability distribution of
three binary variables with high triple interaction. The analysis
also allows us to quantify the amount of redundancy in the mutual
information between pairs of variables, and to assess whether the
information between two variables is or is not mediated by a third
variable. These concepts are applied to the analysis of written
texts. We find that the probability that a given word is found in
a particular location within the text is not only modulated by the
presence or absence of other nearby words, but also, on the
presence or absence of nearby pairs of words. We identify the
words enclosing the key semantic concepts of the text, the
triplets of words with high pairwise and triple interactions, and
the words that mediate the pairwise interactions between other
words.
\end{abstract}

\maketitle


\section{\label{sec:intro}Introduction}

Imagine a game where, as you read through a piece of text, you
occasionally come across a blank space representing a removed or
occluded word. Your task is to guess the missing word. This is an
example sentence, ------ your guess. If you were able to replace
the blank space in the previous sentence with ``make'', or
``try'', or some other related word, you have understood the rules
of the game. The task is called {\sl the Cloze test}
\cite{taylor53} and is routinely administered to evaluate language
proficiency, or expertise in a given subject.

The cues available to the player to solve the task can be divided
into two major groups. First, surrounding words restrict the
grammatical function of the missing word, since, for example, a
conjugated verb cannot usually take the place of a noun, nor vice
versa. Second, and assuming that the grammatical function of the
word has already been surmised, semantic information provided by
the surrounding words is typically helpful. That is, the presence
or absence of specific words in the neighborhood of the blank
space affect the probability of each candidate missing word. For
example, if the word {\sl bee} is near the blank space, the
likelihood of {\sl honey} is larger than when {\sl bee} is absent.

In this paper we study the structure of the probabilistic links
between words due to semantic connections. In particular, we aim
at deciding whether binary interactions between words suffice to
describe the structure of dependencies, or whether triple and
higher-order interactions are also relevant: Should we only care
for the presence or absence of specific words in the vicinity of
the blank space, or does the presence or absence of specific {\sl
pairs} (or higher-order combinations) also matter in our ability
to guess the missing word? For example, one would expect that the
presence of the word {\sl cell} would increase the probability of
words as {\sl cytoplasm}, {\sl phone} or {\sl prisoner}. The word
{\sl wax}, in turn, is easily associated with {\sl ear}, {\sl
candle} or {\sl Tussaud}. However, the conjoint presence of {\sl
cell} and {\sl wax} points much more specifically to concepts such
as {\sl bee} or {\sl honey}, and diminish the probability of words
associated with other meanings of {\sl cell} and {\sl wax}.
Combinations of words, therefore, also matter in the creation of
meaning, and context. The question is how relevant this effect is,
and whether the effect of the pair ({\sl cell} + {\sl wax}) is
more, equal or less than the sum of the two individual
contributions (effect of {\sl cell} + effect of {\sl wax}). Here
we develop the mathematical methods to estimate these
contributions quantitatively.

The problem can be framed in more general terms. In any complex
system, the statistical dependence between individual units cannot
always be reduced to a superposition of pairwise interactions.
Triplet, or even higher-order dependencies may arise either
because three or more variables are dynamically linked together,
or because some hidden variables, not accessible to measurement,
are linked to the visible variables through pairwise interactions.

In 2006, Schneidman and coworkers \cite{schneidman06} demonstrated
that, in the vertebrate retina, up to pairwise correlations
between neurons could account for approximately 90\% of all the
statistical dependencies in the joint probability distribution of
the whole population. This finding brought relief to the
scientific community, since an expansion up to the second order
was regarded sufficient to provide an adequate description of the
correlation structure of the full system. As a consequence, not
much effort has been dedicated to the detection and the
characterization of third or higher-order interactions. To our
knowledge, the present work constitutes the first example offering
an exact description of third-order dependencies. We derive the
relevant information-theoretical measures, and then apply them to
actual data.

As a model system, we work with the vast collection of words found
in written language, since this system is likely to embody complex
statistical dependencies between individual words. The
dependencies arise from the syntactic and semantic structures
required to map a network of interwoven thoughts into an ordered
sequence of symbols, namely, words. The projection from the
high-dimensional space of ideas onto the single dimension
represented by time can only be made because language encodes
meaning in word order, and word relations. In particular, if
specific words appear close to each other, they are likely to
construct a context, or a topic. The context is important in
disambiguating among the several meanings that words usually
have. Therefore, language constitutes a
model system where individual units (words) can be expected to
exhibit high-order interactions.

Statistics and information theory have proved to be useful in understanding language structures. Since Zipf's empirical law  \cite{zipf49} on the frequency of words, and the pioneering work of Shannon \cite{shannon51} measuring the entropy of printed English, a whole branch of science has followed these lines \cite{grassberger89,ebeling94,montemurro10}. In recent years, the discipline gained momentum with the availability of large data sources in the internet \cite{petersen12,perc12,gerlach13,gerlach14}.

In this paper we quantify the amount of double and triple
interactions between words of a given text. In addition, by means
of a careful analysis of the structure of pairwise interactions we
distinguish between pairs of variables that interact directly, and
pairs of variables that are only correlated because they both
interact with a third variable. With these goals in mind, we
define and measure dependencies between words using concepts from
information theory \cite{shannon48,jaynes57,cover}, and apply them
in later sections to the analysis of written texts.

\section{\label{sec:triplet}Statistical dependencies among three variables}

\begin{figure}[ht]
    \begin{center}
    \includegraphics[width=8cm, clip=true]{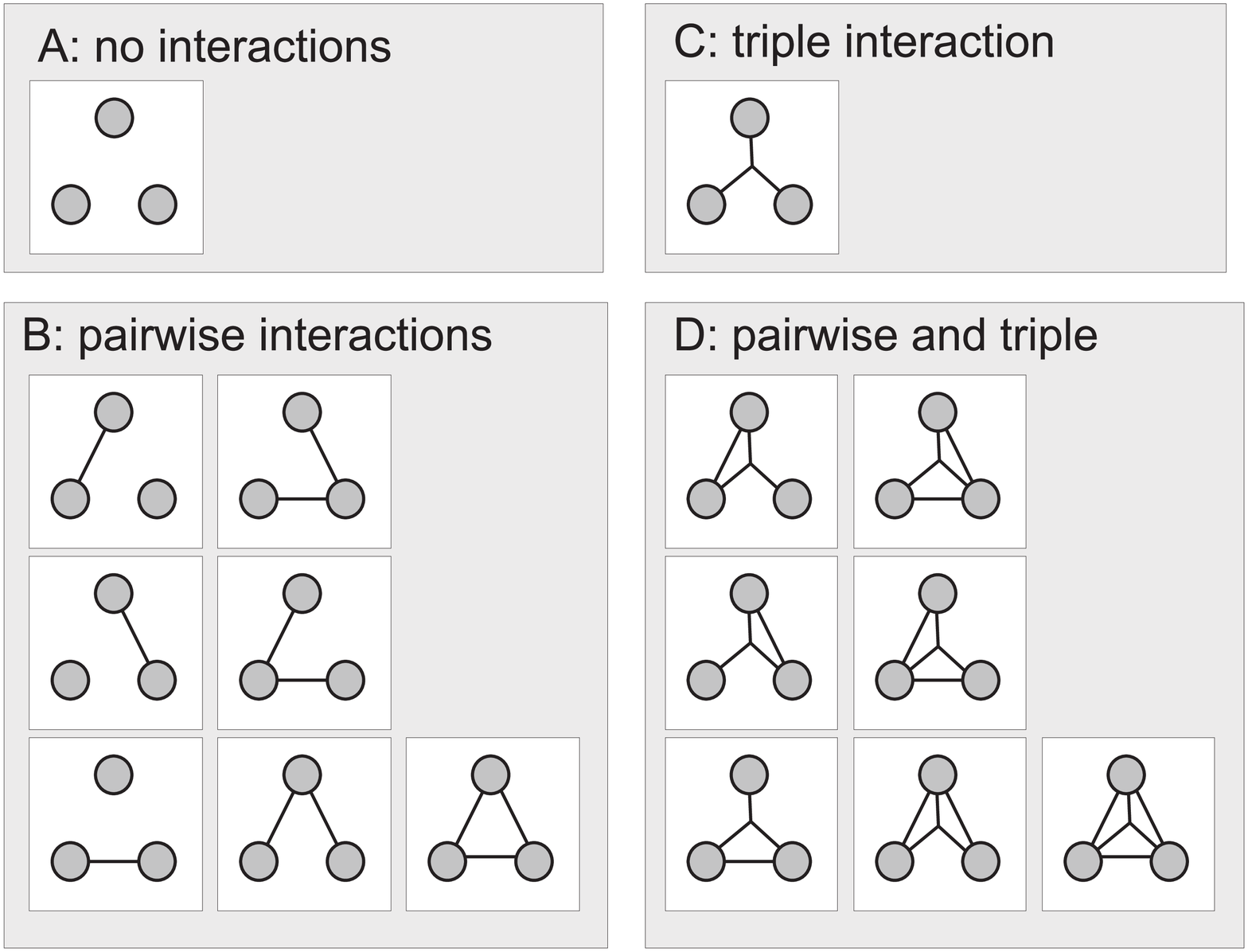}
    \end{center}
\caption{\footnotesize{Different ways in which three
variables may interact. A: The three variables are independent. B:
Only pairwise interactions exist. These may involve 1, 2 or 3
links (from left to right). C: The three variables are connected
by a single triple interaction. D: Double and triple interactions
may coexist. The most general case is illustrated in the
bottom-right panel.}} \label{fig1}
\end{figure}

When it comes to quantifying the amount of statistical dependence
between two variables $X_1$ and $X_2$ with joint probabilities
$p(x_1, x_2)$ and marginal probabilities $p(x_1)$ and $p(x_2)$,
Shannon's mutual information \cite{shannon48, cover}
\begin{equation}
I(X_1; X_2)=\sum_{x_1,x_2} p(x_1, x_2)\log \frac{p(x_1,
x_2)}{p(x_1)p(x_2)} \label{eq:00}
\end{equation}
\noindent stands out for its generality and its simplicity.
Throughout this paper we take all logarithms in base 2, and
therefore measure all information-theoretical quantities in bits.
In Fig.~\ref{fig1}, pairwise statistical dependencies are
represented by the rods connecting two variables (independent
variables appear disconnected). Since $I(X_1; X_2)$ is the
Kullback-Leibler divergence $D[p(x_1, x_2): p(x_1) p(x_2)]$
\cite{cover} between the joint distribution $p(x_1, x_2)$ and its
independent approximation $p(x_1)p(x_2)$, the mutual information
is always non-negative. Moreover, $X_1$ and $X_2$ are independent
if and only if their mutual information vanishes.

Three variables, in turn, may interact in different ways;
Fig.~\ref{fig1} illustrates all the possibilities.
In this section, we discuss several quantities that measure the
strength of the different interactions. So far, no general
consensus has been reached regarding the way in which statistical
dependencies between three variables should be quantified
\cite{darroch62,mcgill54,agresti,martignon00,amari01,bell2003co,schneidman03,nemenman04,vitanyi11,griffith14}.
One attempt in the framework of Information Theory is the
symmetric quantity $I(X_1; X_2; X_3)$, sometimes called the {\sl
co-information} \cite{cover,bell2003co}, defined as
\begin{equation}
\begin{array}{ll}
I(X_1; X_2; X_3) &= I(X_1; X_2) - I(X_1; X_2 | X_3) \\
&= I(X_2; X_3) - I(X_2; X_3 | X_1) \\
&= I(X_3; X_1) - I(X_3; X_1 | X_2),
\end{array}
\label{eq:02}
\end{equation}
where $I(X_i; X_j|X_k)$ is the conditional mutual information,
\begin{equation}
I(X_i; X_j|X_k) = \sum_{x_i, x_j, x_k}p(x_i, x_j,
x_k)\log[\frac{p(x_i, x_j|x_k)}{p(x_i|x_k)p(x_j|x_k)}].
\end{equation}
The co-information measures the way one of the variables (no
matter which) influences the transmission of information between
the other two. Positive or negative values of the co-information
have often been associated with redundancy or synergy between the
three variables, though one should be careful to distinguish
between several possible meanings of the words {\sl synergy} and
{\sl redundancy} (see below, and also
\cite{schneidmanjn,eyherabh10}).

In an attempt to provide a systematic expansion of the different
interaction orders, Amari \cite{amari01} developed an alternative
way of measuring triple and higher-order interactions. His
approach unifies concepts from categorical data analysis and
maximum entropy techniques. The theory is based on a decomposition
of the joint probability distribution as a product of functions,
each factor accounting for the interactions of a specific order.
The first term embodies the independent approximation, the second
term adds all pairwise interactions, subsequent terms orderly
accounting for triplets, quadruplets and so forth. This approach
constitutes the starting point for the present work.

Given the random variables $X_1$, \dots, $X_N$ governed by a joint
probability distribution $p(x_1, \dots, x_n)$, all the marginal
distributions of order $k$ can be calculated by summing the values
of the joint distribution over $n - k$ of the variables. Since
there are $n!/k!(n - k)!$ ways of choosing $n - k$ variables among
the original $n$, the number of marginal distributions of order
$k$ is $n!/k!(n - k)!$ Amari defined the probability distribution
$p^{(k)}(x_1,...,x_N)$ as the one with maximum entropy $H_{\rm
max}^{(k)}$ among all those that are compatible with all the
marginal distributions of order $k$. The maximization of the
entropy under such constraints has a unique solution
\cite{csiszar75}: the distribution allowing variables to vary with
maximal freedom, inasmuch they still obey the restriction imposed
by the marginals. Hence, $p^{(k)}(x_1,...,x_N)$ contains all the
statistical dependencies among groups of $k$ variables that were
present in the original distribution, but none of the dependencies
involving more than $k$ variables.

The interactions of order $k$ are quantified by the decrease of
entropy from $p^{(k-1)}$ to $p^{(k)}$, which can be expressed as a
Kullback-Leibler divergence
\begin{equation}
\begin{array}{ll}
D^{(k)} &= D[p^{(k)}:p^{(k-1)}] \\ \\
&= H_{\rm max}^{(k-1)}-H_{\rm max}^{(k)},
\end{array}
\label{eq:03}
\end{equation}
where $H_{\rm max}^{(k)}$ is the entropy of $p^k$. The last
inequality of Eq.~(\ref{eq:03}) derives from the generalized
Pythagoras theorem \cite{amari01}. As increasing constraints
cannot increase the entropy, $D^{(k)}$ is always non-negative.

The total amount of interactions within a group of $N$ variables,
the so called {\sl multi-information} $\Delta(X_1, \dots, X_N)$
\cite{mcgill54}, is defined as the Kullback-Leibler divergence
between the actual joint probability distribution and the
distribution corresponding to the independent approximation. The
multi-information naturally splits in the sum of the different
interaction orders
\begin{equation}
\begin{array}{ll}
\Delta_{12...N} &= D[p(x_1,...,x_N):p(x_1)...p(x_N)]\\\\
&=\displaystyle \sum_{k=2}^N D^{(k)}.
\end{array}
\label{eq:04}
\end{equation}

For two variables, there are at most pairwise interactions. Their
strength, measured by $D^{(2)}$, coincides with Shannon's mutual
information
\begin{equation}
\begin{array}{ll}
D^{(2)}_{12} &= D[p^{(2)}(x_1, x_2) : p^{(1)}(x_1, x_2)] \\ \\
&= D[p(x_1, x_2): p(x_1) p(x_2)] \\ \\
&= I(X_1; X_2),
\end{array}
\label{eq:05}
\end{equation}
\noindent since the distribution with maximum entropy that is
compatible with the two univariate marginals is $p^{(1)}(x_1, x_2)
= p(x_1) p(x_2)$. This result is easily obtained by searching for
the joint distribution that maximizes the entropy using Lagrange
multipliers for the constraints given by the marginals \citep{kapur89}.

When studying three variables, $X_1$, $X_2$ and $X_3$, we separately
quantify the amount of pairwise and of triple interactions. In this
context, $D^{(3)}_{123}$ measures the amount of statistical
dependency that cannot be explained by pairwise interactions, and is
defined as
\begin{equation}
\begin{array}{ll}
D^{(3)}_{123} &= D[p(x_1, x_2 ,x_3): p^{(2)}(x_1, x_2, x_3)] \\ \\
&= H_{\rm max}^{(2)}-H_{123},
\end{array}
\label{eq:06}
\end{equation}
where $H_{123}$ represents the full entropy of the triplet $H(X_1,
X_2, X_3)$ calculated with $p(x_1, x_2, x_3)$.

The distribution $p^{(2)}(x_1, x_2, x_3)$ contains up to pairwise
interactions. If the actual distribution $p(x_1, x_2, x_3)$
coincides with $p^{(2)}(x_1, x_2, x_3)$, there are no third-order
interactions. Within Amari's framework, hence, if $D^{(3)}_{123} >
0$, some of the statistical dependency among triplets cannot be
explained in terms of pairwise interactions.

Both $I(X_1; X_2; X_3)$ and $D^{(3)}_{123}$ are generalizations of
the mutual information intended to describe the interactions
between three variables, and both of them can be extended to an
arbitrary number of variables \cite{amari01,yeung02}. It is
important to notice, however, that the two quantities have
different meanings. A vanishing co-information ($I(X_1; X_2; X_3)
= 0$) implies that the mutual information between two of the
variables remains unaffected if the value of the third variable is
changed. However, this does not mean that it suffices to measure
only pairs of variables---and thereby obtain the marginals $p(x_1,
x_2), p(x_2, x_3), p(x_3, x_1)$---to reconstruct the full
probability distribution $p(x_1, x_2, x_3)$. Conversely, a
vanishing triple interaction ($D^{(3)}_{123} = 0$) ensures that
pairwise measurements suffice to reconstruct the full joint
distribution. Yet, the value of any of the variables may still
affect how much information is transmitted between the other two.

We shall later need to specify the groups of variables whose
marginals are used as constraints. We therefore introduce a new
notation for the maximum entropy probability distributions and for
the maximum entropies. Let $V$ represent a set of $k$ variables. For
example, if $k = 3$, we may have $V = \{X_1, X_2, X_3\}$. When studying
the dependencies of $k$-th order, we shall be working with all sets
$V_1, \dots, V_r$ that can be formed with $k$ variables, where $r =
n! / k! (n - k)!$ Let $p_{V_1, V_2, \dots, V_r}$ be the probability
distribution of maximum entropy $H_{V_1, V_2, \dots, V_r}$ that
satisfies the marginal restrictions of $V_1, V_2, \dots ,V_k$. Under
this notation,
\begin{equation}
\begin{array}{ll}
p^{(2)}(x_1,x_2,x_3) &=p_{12,13,23}\\\\
p^{(1)}(x_1,x_2,x_3) &=p_{1,2,3}.
\end{array}
\label{eq:07}
\end{equation}
\noindent Respectively, the maximum entropies are $H_{12,13,23}$
and $H_{1,2,3} = H(X_1) + H(X_2) + H(X_3)$. Under the present
notation, the mutual information $I(X_i; X_j)$ is $I_{ij}$, and
the co-information of three variables $X_1, X_2, X_3$ is written
as $I_{123}$.

The amount of pairwise interactions $D^{(2)}_{ij}$ between
variables $i$ and $j$ is known to be bounded by \cite{cover}
\begin{equation}
D^{(2)}_{ij} = I_{ij} \le {\rm min}(H_i, H_j). \label{eq:07a}
\end{equation}
We have derived an analogous bound for triple interactions
(see Appendix \ref{Apendicematematico}). The resulting inequality
links the amount of triple interactions $D^{(3)}_{123}$ with
the co-information $I_{123}$,
\begin{equation} \label{eq:cota}
D^{(3)}_{123} \le  {\rm min}\{I_{12}, I_{23}, I_{31}\} - I_{123}
\le  {\rm min}\{H_1, H_2, H_3\}.
\end{equation}
These bounds imply that pure triple interactions, appearing in the
absence of pairwise interactions (see Fig.~\ref{fig1}C), may only
exist if the co-information $I_{123}$ is negative.

\subsection{\label{ssu:triplet} Characterization of the joint
probability distribution of variables with high triple interactions
}

\label{sec:xor}

\begin{figure}[ht]
    \begin{center}
    \includegraphics[width=8cm]{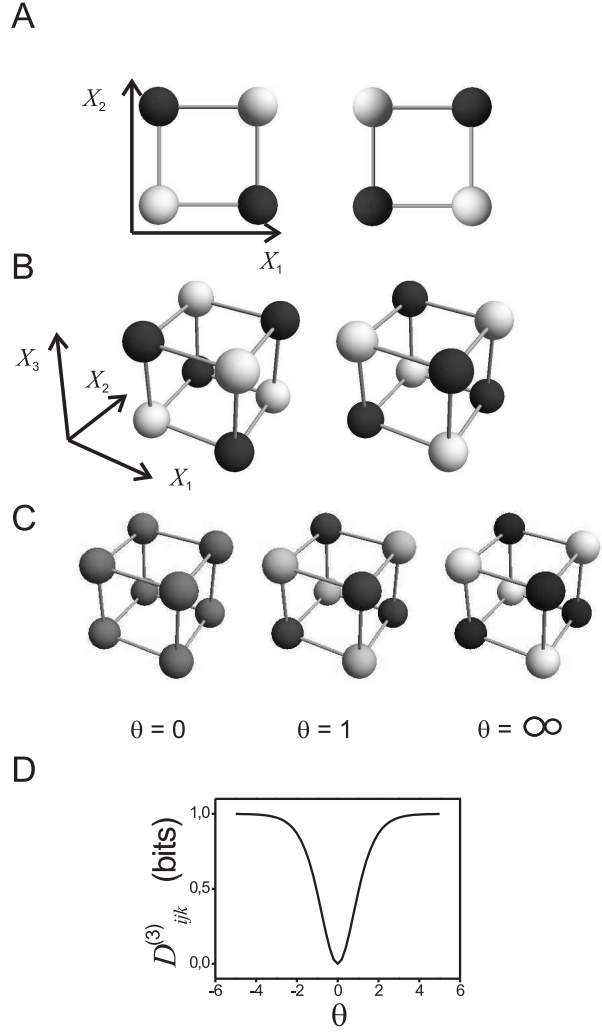}
    \end{center}
\caption{\footnotesize{A: Density plot of the two bivariate
probability distributions that have $I = 1$ bit. Dark states have
zero probability, and white states have $p(x_1, x_2) = 1/2$. B:
Density plot of the two trivariate probability distributions with
$D^{(3)}_{ijk} = 1$ bit. Dark states have zero probability, and
white states have $p(x_1, x_2, x_3) = 1/4$. C: Gradual change
between a uniform distribution and a $XOR$ distribution, for
different values of $\theta$ (Eq.~(\ref{eq:11b})). D: Amount of
triple interactions as a function of the parameter $\theta$. }}
\label{fig:new}
\end{figure}

Two binary variables $X_1$ and $X_2$ can have maximal mutual
information $I_{12} =$ 1 bit in two different situations. For the
sake of concreteness, assume that $X_i = \pm 1$. Maximal mutual
information is obtained either when $X_1 = X_2$ or when $X_1 = -
X_2$. In other words, the joint probability distribution must
either vanish when the two variables are equal, or when the two
variables are different, as illustrated in Fig.~\ref{fig:new}A.
If the mutual information is high, though perhaps not maximal,
then the two variables must still remain somewhat correlated, or
anti-correlated. The joint probability distribution, hence, must
drop for those states where the variables are equal - or
different. In this section we develop an equivalent intuitive
picture of the joint probability distribution of triplets with
maximal (or, less ambitiously, just high) triple interaction.

Consider three binary variables $X_1, X_2, X_3$ taking values $\pm 1$
with joint probability distribution
\begin{equation}
p(x_1, x_2, x_3)= \left\{
\begin{array}{lll}
1/4 &\text{ if } & x_1 x_2 x_3 = -1 \\ \\
0 &\text{ if } & x_1 x_2 x_3 = 1.
\end{array}
\right.
\label{eq:10}
\end{equation}
as illustrated in Fig.~\ref{fig:new}B, left side.
For this probability distribution, the three univariate marginals
$p_1, p_2, p_3$ are uniform, that is, $p_i(1) = p_i(-1) = 1/2$.
Moreover, the three bivariate marginals $p_{12}, p_{23}, p_{31}$
are also uniform: $p_{ij}(1, 1) = p_{ij}(1, -1) = p_{ij}(-1, 1) =
p_{ij}(-1, -1) = 1/4$. The full distribution, however, is far from
uniform, since only half of the 8 possible states have non-vanishing
probability.

The probability distribution of Eq.~(\ref{eq:10}) is henceforth
called a $XOR$ distribution. The name is inspired by the fact that
two independent binary variables $X_1$ and $X_2$ can be combined
into a third dependent variable $X_3 = X_1 \ XOR \ X_2$, where
$XOR$ represents the logical function exclusive-OR. If the two
input variables have equal probabilities for the two states
$\pm1$, then Eq.~(\ref{eq:10}) describes the joint probability
distribution of the triplet $(X_1, X_2, X_3)$.

The maximum-entropy probability compatible
with uniform bivariate marginals is uniform, $p^{(2)}(x_1, x_2, x_3) = 1/8$.
The amount of triple interactions is therefore
\begin{equation}
\begin{array}{ll}
D^{(3)}_{123} &= H_{12,13,23}-H_{123}\\\\
&= 3 \rm{ bits}-2 \rm{ bits}=1 \text{ bit},
\end{array}
\label{eq:11}
\end{equation}
\noindent and $D^{(3)}_{123}=\Delta_{123}$, i.e. all interactions
are tripletwise and $D^{(3)}_{123}$ reaches the maximum value
allowed for binary variables. Of course, the same amount of triple
interactions is obtained for the complementary probability
distribution (a so-called negative-XOR), for which $p(x_1, x_2, x_3) =
1/4$ when $\prod_i x_i = +1$ (see Fig.~\ref{fig:new}B, right side).

So far we have demonstrated that $XOR$ and $-XOR$ distributions
contain the maximal amount of triple interactions. Amari
\cite{amari01} has proved the reciprocal result: If the amount of
triple interactions is maximal, then the distribution is either
$XOR$ or $-XOR$. We now demonstrate that if the joint distribution
lies somewhere in between a uniform distribution and a $XOR$ (or a
$-XOR$) distribution, then the amount of triple interactions lies
somewhere in between 0 and 1, and the correspondence is monotonic.
To this end, we consider a family of joint probability
distributions parametrized by a constant $\theta$, defined as a
linear combination of a uniform distribution $p_u(x_1, x_2, x_3) =
1/8$ and a $\pm XOR$ distribution,
\begin{equation}
\begin{array}{ll}
p_\theta(x_1, x_2, x_3) &\displaystyle = \frac{1}{8} \left( 1 + x_1 x_2 x_3
\tanh\theta  \right),
\end{array}
\label{eq:11b}
\end{equation}
\noindent where $\theta \in (-\infty, +\infty)$. Varying $\theta$
from zero to $\infty$ shifts the $p(x_1, x_2, x_3)$ from the
uniform distribution $p_u$ to the \textit{XOR} probability of
Eq.~(\ref{eq:10}) (see Fig.~\ref{fig:new}C). Negative $\theta$
values, in turn, shift the distribution to $-XOR$. All the
bivariate marginals of the distribution $p_\alpha(x_i, x_j)$ are
uniform, and equal to 1/4. The maximum-entropy model compatible
with these marginals is the uniform distribution $p_u(x_1, x_2,
x_3) = 1/8$. Hence, the amount of triple interactions is
\begin{equation}
D^{(3)}_{123}(\theta) = \frac{1}{2}\left[(1 + \tanh\theta) \log(1 + \tanh\theta)
+  (1 - \tanh\theta) \log(1 - \tanh\theta) \right].
\end{equation}
As shown in Fig.~\ref{fig:new}D, this function is even, and varies
monotonically in each of the intervals $(-\infty, 0)$ and $(0,
+\infty)$. Therefore, there is a one to one correspondence between
the similarity between the $\pm XOR$ distribution and the amount
of triple interactions. The same result is obtained for arbitrary
binary distributions, as argued in the last paragraph of Appendix
\ref{app:solp2}. As a consequence, we conclude that for binary
variables, the $\pm XOR$ distribution is not just one possible
example distribution with triple interactions, but rather, it is
the {\em only} way in which three binary variables interact in a
tripletwise manner. If bivariate marginals are kept fixed, and
triple interactions are varied, then the joint probability
distribution either gains or loses a $XOR$-like component, as
illustrated in Fig.~\ref{fig:new}C.

\section{\label{sec:pair}Triplet analysis of pairwise interactions}

In a triplet of variables $X_1, X_2, X_3$, three possible binary
interactions can exist, quantified by $I(X_1; X_2)$, $I(X_2; X_3)$
and $I(X_3; X_1)$. In this section we characterize the amount of
overlap between these quantities, we bound their magnitude,
and we learn how to distinguish between reducible
and irreducible interactions.

\subsection{Redundancy among the three mutual informations within a triplet}

In the previous section, we saw that when there are only two
variables $X_1$ and $X_2$, $D^{(2)}_{12}$ coincides with the
mutual information $I(X_1; X_2)$. When there are more than two
variables, $D^{(2)}$ can no longer be equated to a mutual
information, since there are several mutual informations in play,
one way per pair of variables: $I(X_1; X_2), I(X_2; X_3)$, etc. In
this section, we derive a relation between all these quantities
for the case of three interacting variables. The multi-information
of Eq.~(\ref{eq:04}) decomposes into pairwise and triple
interactions,
\begin{equation}
\Delta_{123}=D^{(2)}_{123}+D^{(3)}_{123}, \label{eq:12}
\end{equation}
from where we arrive at
\begin{equation}
\begin{array}{ll}
D^{(2)}_{123} &=\Delta_{123}-D^{(3)}_{123}\\\\
&= I_{12}+I_{13}+I_{23}-I_{123}-D^{(3)}_{123}.
\end{array}
\label{eq:13}
\end{equation}
\noindent The total amount of pairwise dependencies, hence, is in
general different from the sum of the three mutual informations.
That is, depending on the sign of $D^{(3)}_{123}+I_{123}$, the
amount of pairwise interactions $D^{(2)}_{123}$ can be larger or
smaller than $\sum I_{ij}$. This range of possibilities
suggests that $\sum I_{ij}- D^{(2)}_{123}$ may be a useful
measure of the amount of redundancy or synergy within the
pairwise interactions inside the triplet, and this is the measure
that we adopt in the present paper.

This measure coincides with the co-information when there are no
triple dependencies, that is, when $D^{(3)}_{123}=0$. In this
case,
\begin{equation}
I_{123}= I_{12}+I_{13}+I_{23}-D^{(2)}_{123}. \label{eq:14}
\end{equation}
\noindent Under these circumstances, a positive value of
$I_{123}$ implies that the
sum of the three mutual informations is larger than the total
amount of pairwise interactions. The content of the three
informations, hence, must somehow overlap. This observation
supports the idea that a positive co-information is associated
with redundancy among the variables. In turn, a negative value of
$I_{123}$ implies that although the maximum entropy distribution
compatible with the pairwise marginals is not equal to $p_1p_2p_3$
(that is, although $D^{(2)}_{123} > 0$), when taken two at a time,
variables do look independent (that is $p_{ij} \approx p_i p_j$).
The statistical dependency between the variables of any pair,
hence, only becomes evident when fixing the third variable. This
behavior supports the idea that a negative co-information is
associated with synergy among the variables.

Of course when $D^{(3)}_{123} > 0$, the co-information is no
longer so simply related to concepts of synergy and redundancy,
not at least, if the latter are understood as the difference
between the sum of the three informations and $D^{(2)}_{123}$.
However, below we show that in actual data, one can often find a
close connection between the amount of triple interactions and the
co-information.

\subsection{Triangular binary interactions}

\label{triangles}

In a group of interacting variables, if $X_1$ has some degree of
statistical dependence with $X_2$, and $X_2$ has some statistical
dependence with $X_3$, one could expect $X_1$ and $X_3$ to show
some kind of statistical interaction, only due to the chained
dependencies $X_1 \to X_2 \to X_3$, even in the absence of a
direct connection. Here we demonstrate that indeed, two strong
chained interactions necessarily imply the presence of a third
connection closing the triangle. In the pictorial representation
of the middle column of Fig.~\ref{fig1}, this means that if only
two connections exist (there is no link closing the triangle),
then the two present interactions cannot be strong. For example,
with binary variables, it is not possible to have $I_{12} = I_{23}
= 1$ bit, and $I_{31} = 0$. The general inequality reads (see the
derivation in Appendix \ref{Apendicematematico})
\begin{equation} \label{eq:triangulitos}
I_{12} + I_{31} - H_1 \le I_{23}.
\end{equation}

\subsection{Identification of pairwise interactions that are mediated through a third variable}

\label{irreducible1}

In the previous section we demonstrated that the chained
dependencies $X_1 \leftrightarrow X_2 \leftrightarrow X_3$ can
induce some statistical dependency between $X_1$ and $X_3$. On the
other hand, it is also possible for $X_1$ and $X_3$ to interact
directly, inheriting their interdependence from no other variable.
These two possible scenarios cannot be disambiguated by just
measuring the mutual information between pairs of variables. In
Appendix \ref{app:irr}, we explain how, starting from the most
general model (illustrated in the lower-right panel of
Fig.~\ref{fig1}), the analysis of triple interactions allows us to
identify those links that can be explained from binary
interactions involving other variables, and those that cannot: the
so-called {\sl irreducible} interactions. Briefly stated, we need
to evaluate whether the interaction between $X_1$ and $X_2$
(captured by the bivariate marginal $p_{12}$) and the interaction
between $X_2$ and $X_3$ (captured by $p_{23}$) suffice to explain
all pairwise interactions within the triplet, including also the
interaction between $X_1$ and $X_3$. To that end, we compute a
measure of the discrepancy between the two corresponding maximum
entropy models,
\begin{equation}
\Delta^{12}_{13,23} = D[p_{12, 13, 23}: p_{13, 32}] = H_{13, 23} -
H_{12, 13, 23} \label{eq:17a}.
\end{equation}
The amount of irreducible interaction, that is, the amount of
binary interaction between $X_1$ and $X_3$ that remains
unexplained through the chain $X_1 \leftrightarrow X_2
\leftrightarrow X_3$ is defined as
\begin{equation}
\Delta^{13} = {\rm min}\left\{I_{12}, \Delta^{12}_{13,23}
\right\}. \label{eq:19a}
\end{equation}
In Sect.~\ref{irreducible2}, we search for pairs of variables with
small irreducible interaction, by computing $\Delta^{13}$ using
all possible candidate variables $X_2$ that may act as mediators.
From them, we keep the one giving minimal irreducible interaction,
that is, the one for which the chain $X_1 \leftrightarrow X_2
\leftrightarrow X_3$ provides the best explanation for the
interaction between $X_1$ and $X_3$.

\section{\label{sec:marg}Marginalization and hidden variables}

\begin{figure}[ht]
    \begin{center}
    \includegraphics[width=5cm, clip=true]{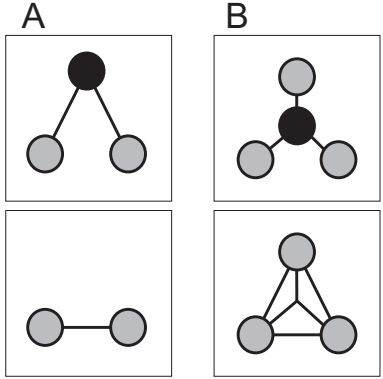}
    \end{center}
\caption{\footnotesize{Examples illustrating the effects of
marginalization in a pair of variables (A) or a triplet (B). In
each case, the variable represented in black drives the other
slave variables, which do not interact directly with each other
(top). However, after marginalizing over the driving variable, a
statistical dependence between the remaining variables appears.
The new interaction can be pairwise (A), or pairwise and
tripletwise (B).}} \label{f:esq}
\end{figure}

Imagine we have a system of $N$ variables that are linked through
just pairwise interactions. In such a system, for any pair of
variables $X_i, X_j$ there is a third variable $X_k$ producing a
vanishing irreducible interaction $\Delta^{ij} = 0$. By selecting
a subset of $k$ variables, we may calculate the $k$-th order
marginal $p^k$, by marginalizing over the remaining $N - k$
variables. As opposed to the original multivariate distribution
$p^N$, the marginal $p^k$ may well contain triple and higher-order
interactions. In other words, there may be pairs of variables
$X_i, X_j$  that belong to the subset for which there is no other
third variable $X_k$ in the subset producing a vanishing
irreducible interaction $\Delta^{ij} = 0$. The high-order
interactions in the subset, therefore, result from the fact that
not all interacting variables are included in the analysis.
Therefore, triple and higher-order statistical dependencies do not
necessarily arise due to irreducible triple and higher-order
interactions: Just pairwise interactions may suffice to induce
them, whenever we marginalize over one or more of the interacting
variables. An example of this effect is derived in
Appendix~\ref{app:example}. In the same way, marginalization may
introduce spurious pairwise interactions between variables that do
not interact directly, as illustrated in Fig.~\ref{f:esq}.
Therefore, even if, by construction, we happen to know that the
system under study can only contain pairwise statistical
dependencies, it may be important to compute triple and
higher-order interactions, whenever one or a few of the relevant
variables are not measured.

Virtually all scientific studies focus their analysis in only a
subset of all the variables that truly interact in the real
system. However, as stated above, neglecting some of the variables
typically induces high-order correlations among the remaining
variables. If such correlations are interpreted within the reduced
framework of the variables under study, they are spurious, at
least, in the sense that there may well be no mechanistic
interaction among the selected variables that gives rise to such
high-order interactions. However, if interpreted in a broader
sense (i.e., a mathematical fact, that may result as a consequence
of marginalization), high-order correlations may be viewed as a
footprint of the marginalized variables, which are often
inaccessible. As such, they constitute an opportunity to
characterize those parts of the system that cannot be described by
the values of the recorded variables.

Below we analyze the statistics of written language. We select a
group of words (each selected word defines one variable), and we
measure the presence or absence of each of these words in
different parts of the book. For simplicity, not all the words in
the book are included in the analysis, so the discarded words
constitute examples of marginalized variables. However,
marginalized variables are not always as concrete as non-analyzed
words. Other non-registered factors may also influence the
presence or absence of specific words, for example, those related
to the thematic topic or the style that the author intended for
each part of the book. These aspects are latent variables that we
do not have access to by simply counting words. An analysis of the
high-order statistics among the subgroup of selected words may
therefore be useful to characterize such latent variables, which
are otherwise inaccessible through automated text analysis.

As an ansatz, we can imagine that each topic affects the
statistics of a subgroup of all the words. The fact that topics
are not included in the analysis is equivalent to having
marginalized over topics. By doing so, we create interactions
within the different subgroups of words. If the topics do not
overlap too much, from the network of the resulting interactions,
we may be able to identify communities of words highly connected,
that are related to certain topics. Variations in the topic can
therefore be diagnosed from variations in the high-order
statistics.

\section{\label{sec:data}Occurrence of words in a book}

Before analyzing a book, all its words are taken in lowercase, and
spaces and punctuation marks are neglected. Each word is replaced
by its base uninflected form using the WordData function from the
program Mathematica\textregistered \cite{worddata}. In this way,
for instance, a word and its plural are considered as the same,
and verb conjugations are unified as well.

In order to construct the network of interactions between words,
we analyze the probability that different words appear near to
each other. The notion of neighborhood is introduced by segmenting
each book into parts. A book containing $M$ words is divided into
$P$ parts, so that there are $M / P$ words per part. We analyze
the statistics of a subgroup of $K$ selected words $w_1, \dots,
w_K$, and define the variables
\begin{equation}
X_i=\left\{
\begin{array}{ll}
1 & \text{ if the word $w_i$ appears in a part}\\\\
-1 & \text{ otherwise}.
\end{array}
\right.
\label{eq:26}
\end{equation}
The different parts of the book constitute the different samples
of the joint probability $p(x_1, x_2, \dots, x_K)$, or of the
corresponding marginals. Notice that if word $w_i$ is found in a
given part of the book, in that sample $X_i = 1$, no matter
whether the word appeared one or many times. The marginal
probability $p(x_i) = (\langle x_i \rangle + 1) / 2$ is the
average frequency with which word $w_i$ appears in one (any) of
the parts. Here, we analyze up to triple dependencies, so we work
with joint distributions of at most three variables $p(x_i, x_j,
x_k)$.

In the present work, we choose to study words that have an
intermediate range of frequencies. We disregard the most frequent
words (which are generally stop words such as articles, pronouns
and so on) because they predominantly play a grammatical role, and
only to a lesser extent they influence the semantic context
\cite{montemurro02}. We also discard the very infrequent words
(those appearing only a few times in the whole book), because
their rarity induces statistical inaccuracies due to limited
sampling \cite{samengo02}. Discarding words implies that only a
seemingly small number of words are analyzed, allowing us to
illustrate the fact that even a small number of variables suffices
to infer important aspects of the structure of the network of
statistical dependencies among words. In other types of data, the
limitation in the number of variables may arise from unavoidable
technical constraints, and not from a matter of choice.

We analyzed two books, \textit{On the Origin of Species} (OS) by
Charles Darwin and \textit{The Analysis of Mind} (AM) by Bertrand
Russell, both taken from Project Gutenberg website
\cite{gutenberg}. Each book was divided into $P = 512$ parts. In
OS, each part contained $295$ words, and in AM, $175$. Parts
should be big enough so that we can still see the structure of
semantic interactions, and yet, the number of parts should not be
too small as to induce inaccuracies due to limited sampling.

In both books, we analyzed $K = 400$ words with intermediate
frequencies. For OS, the analyzed words appeared a total number of
times $n_i$, with $33 \leq n_i \leq 112$. For AM, we analyzed
words with $21 \leq n_i \leq 136$. Since for these words the
number of samples (parts) is much greater than the number of
states (2), entropies were calculated with the maximum likelihood
estimator. We are able to detect differences in entropy of $0.01$
bits, with a significance of $\alpha=0.1 \%$ (see Appendix
\ref{app:sign} for a analysis of significance). A Bayesian
analysis of the estimation error due to finite sampling was also
included, allowing us to bound errors between $0.005$ bits and
$0.01$ bits, depending on the size of the interaction (see
Appendix \ref{app:err}).

\subsection{Statistics of single words}

Before studying interactions between two or more words, we
characterize the statistical properties of single words.
Specifically, we analyze the frequency of individual words, and
their predictability of its presence in one (any) part of the
book. Within the framework of Information Theory, the natural
measure of (un)predictability is entropy.

Using the notation $p_i = p(x_i)$, the entropy $H_i$ is
\begin{equation}
 H_i= -(1-p_i)\log_2(1-p_i)-p_i\log_2 p_i.
\label{eq:27}
\end{equation}
This quantity is maximal ($H = 1$ bit) when $p_i = 1/2$, that is,
when the word $w_i$ appears in half of the parts. When $w_i$
appears in either most of the parts or in almost none, $H_i$
approaches zero. For all the analyzed words, $0 < p_i < 1/2$. In
this range, the entropy $H$ is a monotonic function of $p_i$.

The value of $p_i$, however, is not univocally determined by
the number $n_i$ of times that the word $w_i$ appears in
the book. If $w_i$ appears at most once per part, then
$p_i = n_i/P$. If $w_i$ tends to appear several times per
part, then $p_i < n_i/P$.

In addition, one can determine whether the
fraction of parts containing the word is in accordance with the
expected fraction given the total number of times $n_i$ the word
appears in the whole book. If $n_i$ is half the number of parts
(that is, $n_i = P / 2$), then $p_i = 1/2$ implies that the $n_i$
words are distributed as uniformly as they possibly can: Half of
the parts do not contain the word, and the other half contain it
just once. If, instead, $n_i = 100 P$, a value of $p_i = 1/2$
corresponds to a highly non-uniform distribution: The word is
absent from half of the parts, but it appears many times in the
remaining half.

To formalize these ideas, we compared the entropy of each selected
word with the entropy that would be expected for a word with the
same probability per part $1 / P$, but randomly distributed
throughout the book and sampled $n_i$ times. The binomial
probability of finding the word $k$ times in one (any) part is
\begin{equation}
\hat{p}_i(k)=\frac{n_i!}{k! (n_i - k)!} \
\left(\frac{1}{P}\right)^k \ \left(1 - \frac{1}{P}
\right)^{n_i - k}. \label{eq:28}
\end{equation}
\noindent Equation~(\ref{eq:28}) describes an integer variable. In
order to compare with Eq.~(\ref{eq:27}), we define $Y_i$ as the
binary variable measuring the presence/absence of word $w_i$ in
one (any) part, assuming that the word is binomially distributed.
That is, $Y_i = 0$ if $k = 0$, and $Y_i = 1$ if $k
> 0$. The marginal probability of $p(Y_i = 1)$ is $\hat{p}(k > 0)
= 1-(1 - 1/P)^{n_i}$. This formula is also obtained when all the words in the book are shuffled. In this case $\hat{p}_i(k)$ follows a hypergeometric distribution, such that $\hat{p}_i(k=0)= \binom{M-n_i}{M/P}/ \binom{M}{M/P}= \prod_{j=0}^{n_i-1} (1- \frac{M/P}{M-j}) \cong (1-1/P)^{n_i}$, where the last equality holds when $M \gg n_i$.

\begin{figure}[ht]
\begin{center}
\includegraphics[width=8cm, clip=true]{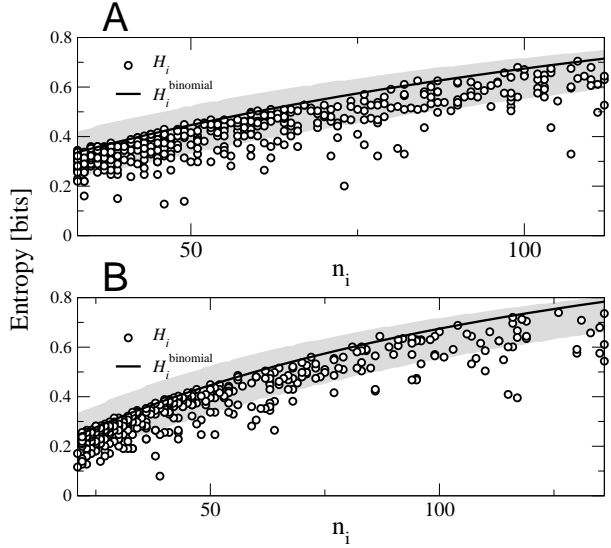}
\end{center}
\caption{\footnotesize{Entropy of the 400 selected words in each
book (one data point per word), compared to the expected entropy
for a binomial variable with the same total count $n_i$
(continuous line), as a function of the total count. Entropies are
calculated with the maximum likelihood estimator. The analytical
expression of Eq.~(\ref{eq:29}) is represented with the black
line, and the gray area corresponds to the percentiles 1\%-99\% of
the dispersion expected in the binomial model, when using a sample
of $n_i$ words. Data points outside the gray area, hence, are
highly unlikely under the binomial hypothesis, even when allowing
for inaccuracies due to limited sampling. A: OS. B: AM.}}
\label{f:ent}
\end{figure}

Hence, the entropy of the binary variable
associated with the binomial (or the shuffled) model is
\begin{equation}
H_i^{\rm binomial}(Y_i)= - (1 - 1/P)^{n_i}\log_2((1 - 1/P)^{n_i})
-(1-(1 - 1/P)^{n_i})\log_2(1-(1 - 1/P)^{n_i}). \label{eq:29}
\end{equation}
The entropy of the variable $X_i$ measured from each book is
compared with the entropy of the binomial-derived variable $Y_i$
in Fig.~\ref{f:ent}.

Even if the process were truly binomial, the estimation of the
entropy may still fluctuate, due to limited sampling. In
Fig.~\ref{f:ent}, the gray region represents the area expected for
98\% of the samples under the binomial hypothesis. We expect 1\%
of the words to fall above this region, and another 1\%, below.
However, in OS, out of 400 words, none of them appears above, and
15\% appear below. In AM, the percentages are 0\% and 16.5\%. In
both cases, the outliers with small entropy are 15 times more
numerous than predicted by the binomial model, and no outliers
with high entropy were found, although 4 were expected for each
book. In both books, hence, individual word entropies were
significantly smaller than predicted by the binomial
approximation, implying that they are not distributed randomly: In
any given part, each word tends to appear many times, or not at
all.

A list of the words with highest difference $(H_i^{\rm
binomial}-H_i)$ is shown in Table \ref{t:01}. Interestingly, most
of these words are nouns, with the first exception appearing in
place 10 (the adjective ``rudimentary'') for OS. As reported
previously \cite{montemurro02}, words with relevant semantic
content are the ones that tend to be most unevenly distributed.

\begin{table}[b]
\caption{\label{t:01} Words with highest difference in entropy
${\Delta H}_i=H_i^{\rm binomial}-H_i$, expressed in bits. Left:
OS. Right: AM. }
\begin{ruledtabular}
\begin{tabular}{ll|ll}
\textrm{Word (OS)}& \textrm{${\Delta H}_i$}&\textrm{Word (AM)}& \textrm{${\Delta H}_i$}\\
\colrule
bee &   0.369   &   proposition &   0.335   \\
cell    &   0.365   &   appearance  &   0.315   \\
slave   &   0.302   &   box &   0.299   \\
stripe  &   0.295   &   datum   &   0.258   \\
pollen  &   0.275   &   animal  &   0.240   \\
sterility   &   0.266   &   objective   &   0.215   \\
pigeon  &   0.252   &   star    &   0.211   \\
fertility   &   0.248   &   content &   0.206   \\
nest    &   0.242   &   emotion &   0.205   \\
rudimentary &   0.234   &   consciousness   &   0.204
\end{tabular}
\end{ruledtabular}
\end{table}

\subsection{Statistics of pairs of words}

In principle, there are two possible scenarios in which the mutual
information between two variables can be high: (a) in each part of
the book the two words either appear together or are both absent,
and (b) the presence of one of the words in a given part excludes
the presence of the other. In Table~\ref{t:01a} we list the pairs
of words with highest mutual information. In all these cases, the
two words in the pair tend to be either simultaneously present or
simultaneously absent (option (a) above).

\begin{table}[b]
\caption{\label{t:01a}Pairs of words with highest mutual
information. Left: OS. Right: AM. The values are in bits.}
\begin{ruledtabular}
\begin{tabular}{lllll|lllll}
\textrm{$w_i$ (OS)} & \textrm{$w_j$} (OS) & \textrm{$I_{ij}$} &
\textrm{$H_i$} & \textrm{$H_j$} & \textrm{$w_i$ (AM)} &
\textrm{$w_j$} (AM) & \textrm{$I_{ij}$} & \textrm{$H_i$} &
\textrm{$H_j$}
\\
\colrule
male & female & 0.242 & 0.504 & 0.409 & 1 & 2 & 0.191 & 0.330 & 0.337   \\
south & america & 0.210 & 0.480 & 0.560 & truth & falsehood & 0.110 & 0.429 & 0.191   \\
reproductive & system & 0.152 & 0.290 & 0.474 & response & accuracy & 0.107 & 0.306 & 0.264   \\
north & america & 0.133 & 0.429 & 0.560 & depend & upon & 0.107 & 0.229 & 0.616   \\
cell & wax & 0.122 & 0.201 & 0.150 & mnemic & phenomena & 0.095 & 0.423 & 0.516   \\
bee & cell & 0.120 & 0.330 & 0.201 & mnemic & causation & 0.090 & 0.423 & 0.381   \\
fertile & sterile & 0.120 & 0.345 & 0.330 & consciousness & conscious & 0.089 & 0.504 & 0.352   \\
deposit & bed & 0.109 & 0.322 & 0.314 & door & window & 0.086 & 0.160 & 0.128   \\
fertility & sterility & 0.109 & 0.352 & 0.322 & stimulus &
response & 0.085 & 0.474 & 0.306 \\
southern & northern & 0.107 & 0.306 & 0.264 & pain & pleasure &
0.079 & 0.171 & 0.181
\end{tabular}
\end{ruledtabular}
\end{table}

The words listed in Table~\ref{t:01a} are semantically related. In
both books, there are examples of words that participate in two
pairs: {\sl cell} is connected to both {\sl bee} and {\sl wax}
(OS) and {\sl mnemic} is connected to both {\sl phenomena} and
{\sl causation} (AM). These examples keep appearing if the lists
of Table~\ref{t:01a} are extended further down. Their structure
corresponds to the double links in the second and third columns of
Figs.~\ref{fig1}B and \ref{fig1}D. As explained in
Sect.~\ref{triangles}, two strong binary links imply that the
third link closing the triangle should also be present. Indeed, in
OS, {\sl america} is linked to both {\sl south} and {\sl north}
(rows 2 and 4 of Table~\ref{t:01a}). The words {\sl south} and
{\sl north} are also linked to each other, but they only appear in
position 32, with a mutual information that is approximately 1/3
of the two principal links. A similar situation is seen with {\sl
bee} and {\sl wax}, both connected to {\sl cell}, although the
direct connection between {\sl bee} and {\sl wax} appears sooner,
in position 16. The same happens in AM with {\sl phenomena} and
{\sl causation}, linked through {\sl mnemic}, which are connected
to each other in the 39th place of the list. These examples pose
the question whether the weakest link in the triangle could be
entirely explained as a consequence of the two stronger links. A
triplet analysis of pairwise interactions allows us to assess
whether such is indeed the case (see Sect.~\ref{irreducible1}).

\begin{figure}[ht]
    \begin{center}
    \includegraphics[width=12.5cm, clip=true]{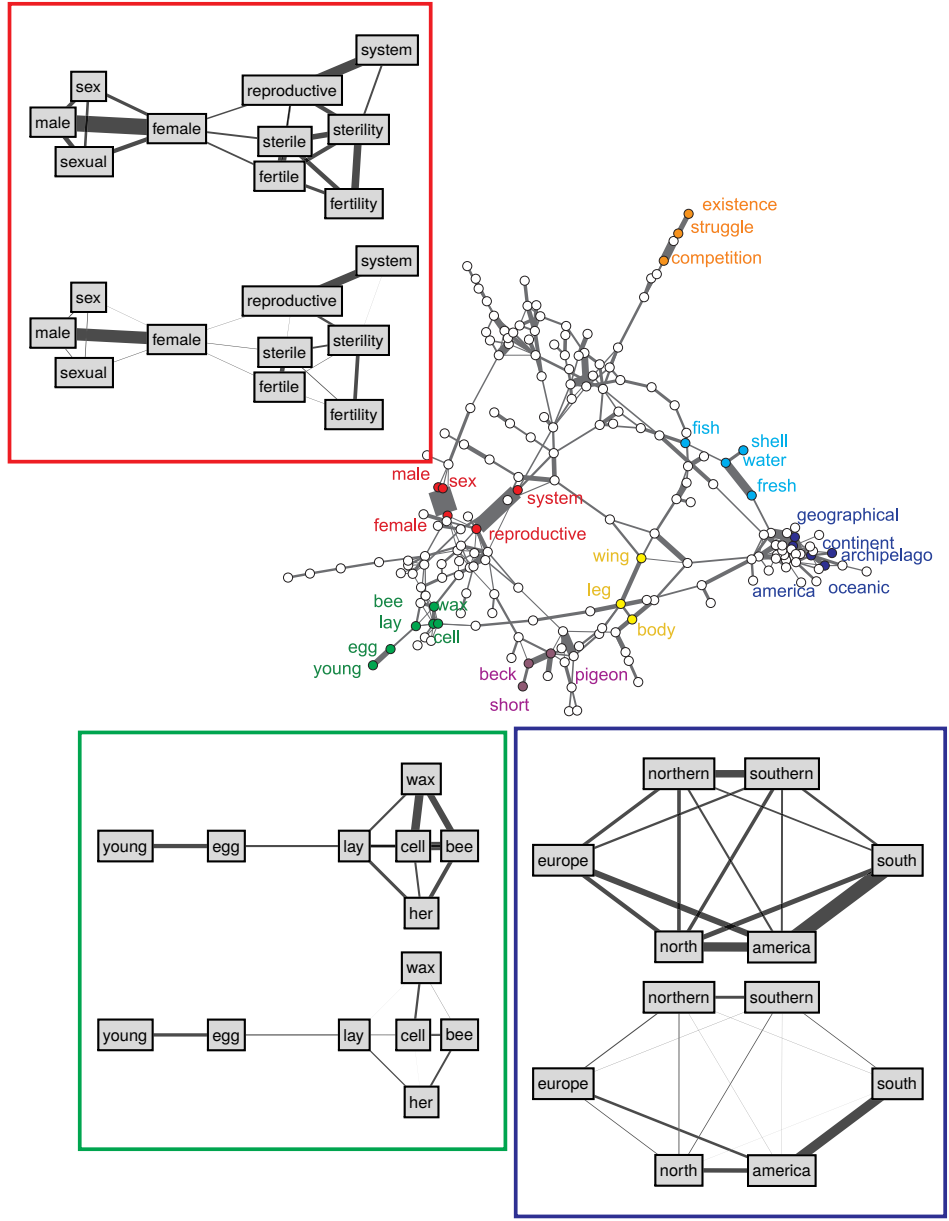}
    \end{center}
\caption{\footnotesize{Central graph: Network of pairwise interactions in OS.
Width of links proportional to the mutual information between
the two connected words. Insets: Detail of selected subnetworks. Top graph: links
proportional to mutual information. Bottom graph: links proportional to
irreducible interaction.}}
\label{f:red}
\end{figure}

We finish the pairwise analysis with a graphical representation of
the words that are most strongly linked with pairwise
connections (left panels of Fig.~\ref{f:red}).
Words belonging to a common topic are displayed in different grey
levels (different colors, online), and tend to form clusters. In each cluster
(insets in Fig.~\ref{f:red}), triplets of words often form triangles of
pairwise interactions. In the central plot, and in the top graph of each inset,
the width of each link is proportional to the mutual information between the
two connected words.

\subsection{Statistics of triplets}

In order to determine whether triple interactions provide a
relevant contribution to the overall dependencies between words,
we compare $D^{(3)}_{ijk}$ with the total amount of pairwise
interactions within the triplet, $D^{(2)}_{ijk}$.

\begin{figure}[ht]
    \begin{center}
    \includegraphics[width=8cm, clip=true]{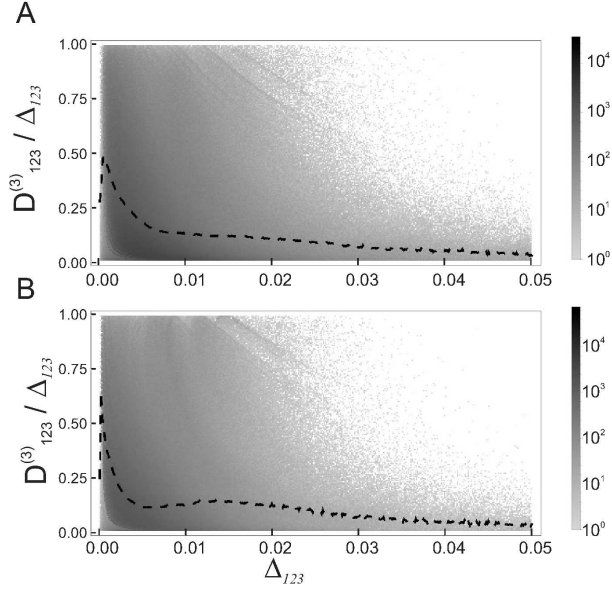}
    \end{center}
\caption{\footnotesize{Fraction of the total interaction within a
triplet $\Delta_{ijk}$ that corresponds to tripletwise
dependencies, $D^{(3)}_{ijk}/\Delta_{ijk}$, as a function of the
total interaction. The grey level of each data point is
proportional to the (logarithm of the) number of triplets at that
location (scale bars on the right). $\Delta_{ijk}$ values above
0.01 bits are significant (see Appendix). A: OS. B: AM.
Dashed line: averages over all triplets with the same $\Delta_{ijk}$.}}
\label{f:bt}
\end{figure}

Figure \ref{f:bt} shows the fraction of the total interaction that
corresponds to triple dependencies, $D^{(3)}_{ijk}/\Delta_{ijk}$,
as a function of the total interaction $\Delta_{ijk}$. The data
extends further to the right, but the triplets with $\Delta_{ijk}
> 0.05$ bits are less than 0.4\%. The first thing to notice is
that the values of the total interaction (values in the horizontal
axis) are approximately an order of magnitude smaller than the
entropies of individual words (see Fig.\ref{f:ent}). Individual
entropies range between 0.1 and 0.9 bits, and interactions are
around 0 and 0.05. In order to get an intuition of the meaning of
such a difference, we notice that if we want to know whether words
$w_i$, $w_j$ and $w_k$ appear in a given part, the number of
binary questions that we need to ask is (depending on the three
chosen words) between 0.3 and 2.7 if we assume the words are
independent ($H_i + H_j + H_k$), and between 0.25 and 2.2, if we
make use of their mutual dependencies ($H_i + H_j + H_k -
\Delta^{(3)}_{123}$). Although sparing $\approx 10\%$ of the
questions may seem a meager gain, it can certainly make a
difference when processing large amounts of data.

The second thing to notice, is that triple interactions are by no
means small as compared to the total interactions within the
triplet, since there are triplets with
$D^{(3)}_{ijk}/\Delta_{ijk}$ of order unity. In other words,
triple interactions are not negligible, when compared to pairwise
interactions. In the triplets with $D^{(3)}_{ijk}/\Delta_{ijk}
\approx 1$, the departure from the independent assumption
resembles the \textit{XOR} behavior (or $-$\textit{XOR}), in the
sense that the states $(x_1, x_2, x_3)$ for which $\prod_i x_i =
1$ have a lower (higher) probability than the states with $\prod_i
x_i = -1$. The first case corresponds to triplets where all pairs
of words tend to appear together, but the three of them are rarely
seen together. In the second case, the words tend to appear either
the three together or each one on its own, but they are rarely
seen in pairs.

\begin{table}[b]
\caption{\label{t:02} Words with highest triple information
$D^{(3)}_{ijk}$. The first column displays a tag that allows us to
identify each triplet in Fig.~\ref{f:d3ci}. The last column
indicates whether the triplet behaves as a $XOR$ gate (+1) or a
$-XOR$ ($-$1). Top: OS. Bottom: AM. Values in bits.}
\begin{ruledtabular}
\begin{tabular}{llllllcc}
Tag&$i$&$j$&$k$&$D^{(3)}_{ijk}$&$I_{ijk}$&$D^{(3)}/\Delta$& $XOR$\\
\colrule
$\alpha$ & america &   south   &   north   &   0.065   &   0.005   &   0.16 & $+1$   \\
$\beta$ & inherit &   occasional  &   appearance  &   0.040   &   $-0.040$  &   0.96  & $
-1$   \\
$\gamma$ & action  &   wide    &   branch  &   0.036   &   $-0.036$  &   0.93  & $-1$   \\
$\delta$ & europe  &   perhaps &   chapter &   0.036   &   $-0.036$  &   0.90 & $-1$    \\
$\epsilon$ & climate &   expect  &   just    &   0.035   &   $-0.035$  &   0.97  & $-1$   \\
\colrule
$\alpha$ & speak   &   causation   &   appropriate &   0.041   &   $-0.041$  &   0.93 & $-1$    \\
$\beta$ & sense   &   perception  &   natural &   0.033   &   $-0.033$  &   0.90 & $-1$    \\
$\gamma$ & since   &   actual  &   wholly  &   0.033   &   $-0.033$  &   0.90  & $-1$   \\
$\delta$ & wish    &   me  &   connection  &   0.033   &   $-0.033$  &   0.95  & $-1$   \\
$\epsilon$ & consist &   should  &   life    &   0.033   &   $-0.033$  &   0.92 & $-1$     \\
\end{tabular}
\end{ruledtabular}
\end{table}

Table \ref{t:02} shows the words with largest triple information.
These interactions are well above the significance threshold of
$0.01$ bits. The triplet (\textit{america}, \textit{south},
\textit{north}) is similar to a $XOR$ gate, so these words tend to
appear in pairs but not all three together. In certain contexts
the author uses the combination \textit{south america}, in other
contexts, \textit{north america}, and yet in others, he discusses
topics that require both \textit{south} and \textit{north} but no
\textit{america}.

Most of the triplets in Table~\ref{t:02} have triple information
values that are equal in magnitude to the co-information but with
opposite sign, that is, $D^{(3)}_{ijk} \approx -I_{ijk}$. Besides,
for these triplets, most of the interaction is tripletwise, that
is, $D^{(3)}_{ijk} / \Delta_{123} \approx 1$.
\begin{figure}[ht]
    \begin{center}
    \includegraphics[width=8cm, clip=true]{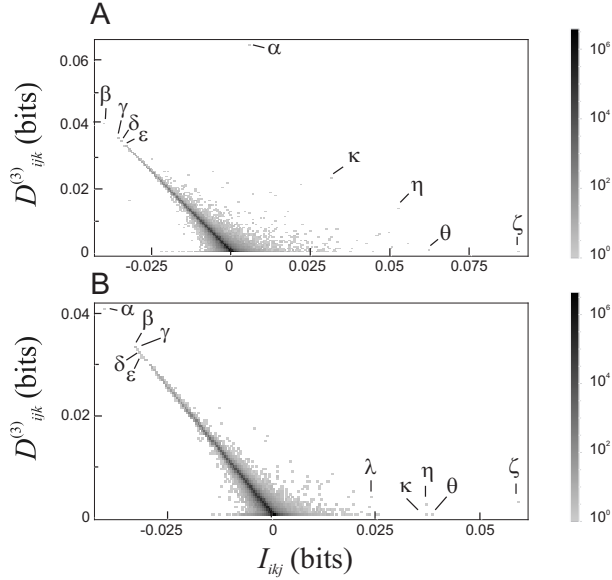}
    \end{center}
\caption{\footnotesize{Triple information $D^{3}_{ijk}$ as a
function of the co-information $I_{ijk}$ for all triplets. The
grey level of each data point is proportional to the (logarithm of
the) number of triplets at that location (scale bars on the
right). $\Delta_{ijk}$ values above 0.01 bits are significant (see
Appendix). A: OS. B: AM.}} \label{f:d3ci}
\end{figure}
To determine whether such tendency is preserved throughout the
population, in Fig.~\ref{f:d3ci} we plot the triple information
$D^{(3)}_{ijk}$ as a function of the co-information $I_{ijk}$ for
all triplets. We see that the vast majority of triplets are
located along the diagonal $D^{(3)}_{ijk} \approx - I_{ijk}$. In
order to understand why this is so, we analyze how data points are
distributed when picking a triplet of words randomly. The cases A,
B, C and D of Fig.~\ref{fig1} are ordered in decreasing
probability. That is, picking three unrelated words
(Fig.~\ref{fig1}A) has higher probability that picking a triplet
with only pairwise interactions (B), which is still more likely
than picking a case with only triple interactions (C), leaving the
case of double and triple interactions (D) as the least probable.
All cases with no triple interaction (A and B) fall on the
horizontal axis $D^{(3)}_{ijk} = 0$ in Fig.~\ref{f:d3ci}.
Therefore, in order to understand why points outside the
horizontal axis cluster along the diagonal we must analyze the
triplets that do have a triple interaction (panels C and D in
Fig.~\ref{fig1}). We begin with case C, because it has a higher
probability than case D. This case corresponds to $D^{(3)}_{ijk}
> 0$ and $I_{ij} = I_{jk} = I_{ki} \approx 0$. It is easy to see that in
these circumstances, $p^{2} \approx p_i p_j p_k$, and hence,
$D^{(3)}_{ijk} \approx -I_{ijk}$. We continue with the left column
of case D, since having a single pairwise interaction has higher
probability than having more. This case corresponds to
$D^{(3)}_{ijk} > 0$, $I_{ij} = I_{jk} \approx 0$ and $I_{ki} > 0$,
for some ordering of the indexes $i, j, k$. In these
circumstances, $p^{2} \approx p_{ij}p_{ik}p_{jk}/p_i p_j p_k$,
which again implies that $D^{(3)}_{ijk} \approx -I_{ijk}$.
Therefore, all triplets containing some triple interaction and at
most a single pairwise interaction fall along the diagonal in
Fig.~\ref{f:d3ci}. The only outliers are triplets with
$D^{(3)}_{ijk} > 0$ and at least two links with pairwise
interactions, which, as derived in Sect.~\ref{triangles}, most
likely contain also the third pairwise link. Such highly connected
triplets are typically few.

From Eq.~(\ref{eq:13}) we see that the triplets that are near the
diagonal are neither synergistic nor redundant, that is, $I_{ij} +
I_{jk} + I_{ki} \approx D^{(2)}_{ijk}$. Those located above the
diagonal have redundant pairwise information ( $I_{ij} + I_{jk} +
I_{ki} > D^{(2)}_{ijk}$), whereas those below are synergistic. In
the two analyzed books, very few ($\approx 10 $) triplets satisfy
$\sum I_{ij} - D^{(2)} < -0.01$ bits. Contrastingly, $\approx 300$
triplets have significant redundant pairwise information ($\sum
I_{ij} - D^{(2)} > 0.01$ bits). The triplets located far from the
diagonal correspond, in both cases, to those with a large total
dependency ($\Delta \gtrsim 0.1$ bits). Table \ref{t:03} displays
the words with highest redundant pairwise interaction, that is,
$I_{ij} + I_{jk} + I_{ki} - D^{(2)}_{ijk}$.
\begin{table}[b]
\caption{\label{t:03} Triplets with highest redundant pairwise
information $D^{(3)}_{ijk} + I_{ijk} = I_{ij} + I_{jk} + I_{ki} -
D^{(2)}_{ijk}$. The first column displays a tag that allows us to
identify each triplet in Fig.~\ref{f:d3ci}. Top: OS. Bottom:
AM. Values in bits.}
\begin{ruledtabular}
\begin{tabular}{lllll}
Tag & $i$&$j$&$k$&$D^{(3)}_{ijk}+I_{ijk}$\\
\colrule
$\zeta$ & bee &   cell    &   wax &   0.089   \\
$\alpha$ & america &   south   &   north   &   0.070   \\
$\eta$ & glacial &   southern    &   northern    &   0.065   \\
$\theta$ & mountain    &   glacial &   northern    &   0.062   \\
$\kappa$ & male    &   female  &   sexual  &   0.057   \\
\colrule
$\zeta$ & leave   &   door    &   window  &   0.061   \\
$\eta$ & stimulus    &   response    &   accuracy    &   0.039   \\
$\theta$ & mnemic  &   phenomena   &   causation   &   0.038   \\
$\kappa$ & truth   &   false   &   falsehood   &   0.036   \\
$\lambda$ & place   &   2   &   1   &   0.027   \\
\end{tabular}
\end{ruledtabular}
\end{table}
With the exception of data point $\alpha$ (\textit{america},
\textit{south}, \textit{north}), the triplets that have highest
redundancy tend to be in the lower right part of
Fig.~\ref{f:d3ci}, whereas the ones with highest triple
interaction lie in the upper left corner.

\subsection{Identification of irreducible binary interactions}
\label{irreducible2}

Using the tools of Sect.\ref{irreducible1}, here we identify the
pairs of words that interact only because the two of them have
strong binary interactions with a third word. In the first place,
the pairs of words whose mutual information is larger than the
significance level (0.01 bits) are selected. For those pairs, the
irreducible interaction is calculated by considering all other
candidate intermediary words, and selecting the one that minimizes
Eq.~(\ref{eq:19a}). We observe that many pairs have a low
irreducible interaction, implying that their dependency can be
understood by a path that goes through a third variable $X_k$,
such as
\begin{equation}
\displaystyle p(x_i,x_j) \approx \sum_{x_k} \frac{p(x_i,x_k)p(x_k,x_j)}{p(x_k)}.
\label{eq:33}
\end{equation}
\noindent In these situations, the behavior of the pair $\{X_i,
X_j\}$ can be predicted from the dependency between $\{X_i, X_k\}$
and the dependency between $\{X_k, X_j\}$.

In Table \ref{t:06}, we list the pairs $(i, j)$ of words that have
smallest irreducible interaction, including the third word ($k$)
that acts as a mediator.
\begin{table}[b]
\caption{\label{t:06} Pairs of words with lowest irreducible
interaction. The first column displays a tag that allows us to
identify each triplet in Fig.~\ref{f:d3ci}. Top: OS. Bottom:
AM. Values in bits.}
\begin{ruledtabular}
\begin{tabular}{lllccl}
&$i$&$j$&$I_{ij}$&$\Delta^{ij}$&$k_{med}$\\
\colrule
$\zeta$ & bee &   wax &   0.093   &   0.003   &   cell    \\
$\alpha$ & south   &   north   &   0.071   &   0.001   &   america \\
$\lambda$ & continent   &   south   &   0.032   &   0.001   &   america \\
$\mu$ & lay &   wax &   0.032   &   0.000   &   cell    \\
$\nu$ & southern    &   arctic  &   0.031   &   0.001   &   northern    \\
\colrule
$\theta$ & phenomena   &   causation   &   0.042   &   0.004   &   mnemic  \\
$\eta$ & stimulus    &   accuracy    &   0.039   &   0.000   &   response    \\
$\lambda$ & place   &   2   &   0.028   &   0.000   &   1   \\
$\mu$ & proposition &   falsehood   &   0.024   &   0.002   &   truth   \\
$\nu$ & proposition &   door    &   0.022   &   0.000   &   window  \\
\end{tabular}
\end{ruledtabular}
\end{table}
In these triplets, most of the interaction between words $w_i$ and
$w_j$ is explained in terms of $w_k$. Mediators tend to have a
high semantic content, and to provide a context in which the other
two words interact. Besides, the triplets $(i, j, k)$ in
Table~\ref{t:06} tend to cluster in the lower right corner of
Fig.~\ref{f:d3ci}, implying that pairs of words share redundant
mutual information.

The number of pairs with significant mutual information (i.e.,
$I_{ij}>0.01$ bits), and whose interaction is explained at least
in a $90\%$ through a third word (i.e., $\Delta^{ij} / I_{ij} <
0.1$) is higher in the book OS ($108$) than in book AM ($19$).
Out of the $108$ pairs of OS, $16$ are explained through
the word \textit{cell}, $12$ through \textit{america}, $8$ through
\textit{northern}, $6$ through \textit{glacial}, $6$ through
\textit{sterility} and so on. The fact that specific words
tend to mediate the interaction between many pairs suggests
that they may act as hubs in the network.

In the right panels of Fig.~\ref{f:red}, we see the network of
irreducible interactions. When compared with the network of mutual
informations (left panels), the irreducible network contains
weaker bonds, as expected, since by definition, $\Delta_{ij}$ cannot
be larger than $I_{ij}$. In the figure, we can identify some of the pairs
of Table~\ref{t:06}, whose interaction is mediated by a third word.
Such pairs appear with a significantly weaker bond
in the right panel, as for example,  \textit{bee}-\textit{wax}
(mediator = \textit{cell}, OS), and
\textit{stimulus}-\textit{accuracy}, (mediator = \textit{response}, AM).
Moreover, one can also identify the pairs whose interaction is
intrinsic (that is, not mediated by a third word) as those where the link on the right
has approximately the same width as on the left. Notable examples
are \textit{male}-\textit{female} (OS), and  \textit{depend}-\textit{upon}.

\section{\label{sec:con}Conclusions}

In this paper, we developed the information-theoretical tools to
study triple dependencies between variables, and applied them to
the analysis of written texts. Previous studies had proposed two
different measures to quantify the amount of triple dependencies:
the co-information $I_{ijk}$ and the total amount of triple
interactions $D^{(3)}$. Given that there is a certain controversy
regarding which of these measures should be used, it is important
to notice that $I_{ijk}$ is a function of three specific variables
$X_1, X_2, X_3$, whereas $D^{(3)}$ is a global measure of all triple
interactions within a wider set of $N$ variables, with $N \ge 3$.
Therefore, it only makes sense to compare the two measures when
$D^{(3)}$ is calculated for the same group of variables as
$I_{ijk}$, which implies using $N = 3$.

The two measures have different meanings. Whereas the
co-information quantifies the effect of one (any) variable in the
information transmission between the other two, the amount of
triple interactions measures the increase in entropy that results
from approximating the true distribution $p_{ijk}$ by the
maximum-entropy distribution that only contains up to pairwise
interactions. When studied with all generality, these two
quantities need not be related, that is, by fixing one of them,
one cannot predict the value of the other. When restricting
the analysis to binary variables, however, a link between them
arises. Three binary variables are characterized by a probability
distribution over $2^3$ possible states. Due to the
normalization restriction, the distribution is determined once the
probability of 7 states are fixed. Choosing those 7 numbers
is equivalent to choosing the three entropies $H_i, H_j, H_k$, the
three mutual informations $I_{ij}, I_{jk}, I_{ki}$, and one more
parameter. This extra parameter can be either the co-information
$I_{ijk}$ (in which case the triple interaction $D^{(3)}$ is
fixed), or the triple
interaction $D^{(3)}$ (in which case the co-information $I_{ijk}$
is fixed). Hence, although in general the co-information and the
amount of triple interactions are not related to one another, for
binary variables, once the single entropies and the pairwise
interactions are determined, $I_{ijk}$ and $D^{(3)}$ become linked.
In this particular situation, hence, there is no
controversy between the two quantities, because they both
provide the same information, only with different scales.

Moreover, we have shown that when pooling together
all the triplets in the system, and now without fixating the value of
individual entropies or pairwise interactions, $I_{ijk}$ and
$D^{(3)}$ often add up to zero. This effect results from the fact
that most triplets contain at most a single pairwise interaction.
Hence, for most of the triplets the two measures
provide roughly the same information. The exception involves the
triplets containing at least two binary interactions, which are
likely to contain all three interactions, in view of
Sect.~\ref{triangles}.

One could repeat the whole analysis presented here, but with $X_i$
= number of times the word appeared in a given part (instead of
the binary variable \textit{appeared} / \textit{not appeared}).
This choice would transform the binary approach into an integer
description, which could potentially be more accurate, if enough
data are available. It should be borne in mind, however, that the
size of the space grows with the cube of the number of states, so
serious undersampling problems are likely to appear in most real
applications. We choose here the binary description to ensure good
statistics. In addition, this choice allowed us to  (a) relate
triple interactions with the $\pm XOR$ gate, and (b) related the
co-information with the amount of triple interactions.

In the present work we studied interactions between words in
written language through a triple analysis. This approach allowed
to accomplish two goals. First, we detected pure triple
dependencies that would not be detectable by studying pairs of
variables. Second, we determined whether pairwise interactions can
be explained through a third word.

We found that on average, 11\% and 13\% of the total interaction
within a group of three words is pure tripletwise. On average,
triple dependencies are weaker than pairwise interactions.
However, in 7\% and 9\% of the total number of triplets, triple
interactions are larger than pairwise. Although this is a small
fraction of all the triplets, all the 400 selected words
participate in at least one such triplet. Hence, if word
interactions are to be used to improve the performance in a Cloze
test, triple interactions are by no means negligible.

We believe that in particular for written language the presence of
triple interactions is mainly due the marginalization over the
latent topics. For example, the triplet (\textit{america},
\textit{south}, \textit{north}) resembles a $XOR$ gate, so
variables tend to appear two at a time, but not alone, nor the
three together. Imagine we include an extra variable (this time, a
non-binary variable), specifying the geographic location of the
phenomena described in each part of the book. The new variable
would take one value in those parts where Darwin describes events
of North America, another value for South America, and yet other
values in other parts of the globe. If these topic-like variables
are included in the analysis, the amount of high order
interactions between words is likely to diminish, because complex
word interactions would be mediated by pairwise interactions
between words and topics. However, since topic-like variables are
not easily amenable to automatic analysis, here we have restricted
the study to word-like variables. We conclude that high-order
interactions between words is likely to be the footprint of having
ignored (marginalized) over topic-like variables.

\begin{acknowledgments}
We thank Agencia Nacional de Promoci\'on Cient\'{\i}fica y
Tecnol\'ogica, Comisi\'on Nacional de Energ\'{\i}a At\'omica and
Universidad Nacional de Cuyo for supporting the present research.
\end{acknowledgments}

\appendix

\section{\label{Apendicematematico} Mathematical proofs}

\subsubsection{Derivation of the bound in Eq.~(\ref{eq:cota})}

As imposing more restrictions cannot increase the entropy, $H_{12,
23, 31} \leq H_{12, 23}$. Using the fact that $H_{12, 23} = H_{12}
+ H_{23} - H_{2}$ (see Appendix~\ref{app:solp2}), it follows from
Eq.~(\ref{eq:06}) that
\begin{equation}
\begin{array}{ll}
D^{(3)}_{123} &\leq H_{12,23}-H_{123}\\\\
D^{(3)}_{123} &\leq I_{13|2}.
\end{array}
\label{eq:08}
\end{equation}
\noindent This inequality is tight, since a probability distribution
exists for which the equality is fulfilled: when $H_{12, 23} =
H_{12, 23, 31}$, that is, when $p_{12, 23, 31}(x_1, x_2, x_3) =
p_{12} \ p_{23} / p_2$.

The derivation can be done removing any of the restrictions
$V\in\{12,13,23\}$. Therefore,
\begin{equation}
\begin{array}{ll}
D^{(3)}_{123} &\leq \min\{I_{12|3}, I_{23|1}, I_{13|2}\}\\\\
D^{(3)}_{123} &\leq \min\{I_{12},I_{13},I_{23}\}-I_{123},
\end{array}
\label{eq:09}
\end{equation}
\noindent where $I_{123}$ is the co-information. From
Eq.~(\ref{eq:09}), it also follows that
\begin{equation}
D^{(3)}_{123} \leq \min\{H_{1},H_{2},H_{3}\}. \label{eq:9a}
\end{equation}

\subsubsection{Derivation of Eq.~(\ref{eq:triangulitos})}

Inserting the upper bound of Eq.~(\ref{eq:08}) in
Eq.~(\ref{eq:13}),
\begin{eqnarray}
I_{12} + I_{23} + I_{31} &=& I_{123} + D^{(2)}_{123} +
D^{(3)}_{123} \nonumber \\
&\le& I_{123} + D^{(2)}_{123} + I_{23|1} \nonumber \\
&=& I_{23} - \cancel{I_{23|1}} + D^{(2)}_{123} +
\cancel{I_{23|1}}.
\end{eqnarray}
Therefore,
\begin{equation}
I_{12} + I_{31} \le D^{(2)}_{123}. \label{e1}
\end{equation}
In addition, since reducing the number of marginal restrictions
cannot diminish the entropy of the maximum entropy distribution,
\begin{eqnarray}
D^{(2)}_{123} &=& - H[p_{12,23,31}] + H_1 + H_2 + H_3 \nonumber \\
&\le& -H[p_{23}] + H_1 + H_2 + H_3 \nonumber \\
&=& I_{23} + H_1. \label{e2}
\end{eqnarray}
Combining Eqs.~(\ref{e1}) and (\ref{e2}),
\[ I_{12} + I_{31} - H_1 \le I_{23}.\]
Therefore, if $I_{12}$ and $I_{31}$ are large, $I_{23}$ cannot be
too small.

\section{\label{app:solp2}Maximum entropy solution}

The problem of finding the probability distribution that maximizes
the entropy under linear constrains, such as fixing some of the
marginals, has a unique solution \cite{csiszar75}. Although no
explicit closed form is known for the case where each variable
varies in an arbitrary domain, there are procedures, for example
the iterative proportional fitting \cite{csiszar75}, that converge
to the solution.

In some special cases a closed form exists. For example, when the
univariate marginals are fixed, the solution is the product of
such marginals. Another case is when we look for the maximum
entropy distribution of three variables $\hat{p}(x_1, x_2, x_3)$
that satisfies two constraints---for example $p(x_1, x_2)$ and
$p(x_2, x_3)$---out of the three bivariate marginals. Posing the
maximization problem through Lagrange multipliers, we obtain a
solution of the form
\begin{equation}
\displaystyle \hat{p}(x_1, x_2, x_3) = f_1(x_1, x_2)f_2(x_2, x_3).
\label{ea0:00}
\end{equation}
If we enforce the marginal constrains and the normalization, we
get
\begin{equation}
\displaystyle \hat{p}(x_1, x_2, x_3) = \frac{p(x_1, x_2) p(x_2,
x_3)} {p(x_2)}, \label{ea0:01}
\end{equation}
\noindent which is known as the pairwise approximation. The
entropy of this distribution is
\begin{equation}
\displaystyle H[\hat{p}] = H_{12, 23} = H_{12} + H_{23} - H_2.
\label{ea0:011}
\end{equation}

Below we derive the solution $p^{(2)}(x_1, x_2, x_3)$ in the
special case of three binary variables ($X_i = \pm 1$). This
solution has maximum entropy and satisfies the three second order
marginal constrains, $p(x_1, x_2)$, $p(x_1, x_2)$ and $p(x_2,
x_3)$. In principle, eight variables need to be determined, one
for the probability of each state. However, considering the
normalization condition, the constraints on the three univariate
marginals, and on the three bivariate marginals, we are left with
only a single free variable. As shown in previous studies
\cite{amari01,martignon00}, the problem reduces to finding the
root of a cubic equation. Since we are interested in comparing
this solution with the joint probability $p(x_1, x_2, x_3)$, a
convenient and conceptually enlightening way of expressing the
solution $p^{(2)}(x_1, x_2, x_3)$, as in the work of Martignon
\cite{martignon00}, is
\begin{equation}
p^{(2)}(x_1,x_2,x_3)=p(x_1,x_2,x_3)-\delta\prod_i x_i,
\label{ea0:02}
\end{equation}
\noindent where the value of $\delta$ is such that the
probabilities remain in the simplex, that is, $p^{(2)}(\textbf{x})
\in [0,1]$. For the marginals, we get
\begin{equation}
\begin{array}{ll}
p^{(2)}(x_i,x_j) &=p^{(2)}(x_i,x_j,1)+p^{(2)}(x_i,x_j,-1)\\
&=p(x_i,x_j,1)+p(x_i,x_j,-1)-\delta+\delta \\
&=p(x_i,x_j).
\end{array}
\label{ea0:03}
\end{equation}
\noindent The value of $\delta$ is obtained from
\begin{equation}
\begin{array}{l}
\displaystyle \prod_{\textbf{x}/ \prod_i x_i=1} p^{(2)}(x_1,x_2,x_3)  =\prod_{\textbf{x}/ \prod_i x_i=-1} p^{(2)}(x_1,x_2,x_3),
\end{array}
\label{ea0:04}
\end{equation}
\noindent condition ensuring that the coefficient accounting for
the triple interaction in the log-linear model vanishes
\cite{amari01}. Eq.~(\ref{ea0:04}) reduces to the previously
mentioned cubic equation on $\delta$.

If the solution is $\delta=0$, then the probability $p$ is the one
with maximum entropy. Otherwise, the probability $p$ departs from
$p^{(2)}$, implying that, up to a certain degree, the multivariate
distribution resembles either the \textit{XOR} gate, or its
opposite.

We close this section by discussing the effect of varying the
amount of triple interactions while keeping all bivariate
marginals fixed, as discussed in Sect.~\ref{sec:xor}. There we
proved that when $p(x_1, x_2, x_3)$ took the shape of
Eq.~(\ref{eq:11b}), then the amount of triplet interactions was a
measure of the similarity between the joint distribution and a
$\pm XOR$ distribution. Here we extend this result to arbitrary
distributions. We have demonstrated here that $p(x_1, x_2, x_3)$
can always be written as $p(x_1, x_2, x_3) \propto p^{(2)}(x_1,
x_2, x_3) + \delta x_1 x_2 x_3$, where $p^{(2)} (x_1, x_2, x_3)$
is the maximum entropy model compatible with the bivariate
marginals of the original distribution, and $\delta$ is a certain
constant. Amari showed that if $\delta = 0$, there are no triple
interactions. Pushing his argument further, here we notice that if
the bivariate marginals are kept fixed, the only way of changing
the amount of triple interactions is to vary the value of
$\delta$. The size of $\delta$ determines the degree of similarity
between $p(x_1, x_2, x_3)$ and a $\pm XOR$ distribution.
Therefore, once the bivariate marginals are fixed, the only
parameter that can be manipulated in order to change the amount of
triple interactions is the one that quantifies the size of the
$\pm XOR$ component.

\section{Irreducible interactions}

\label{app:irr}

Following the ideas from \cite{nemenman04,margolin10}, we wish to
detect whether the statistical dependencies among a group of
variables $V = \{X_1, \dots, X_k\}$ contain all possible
interactions, or whether some of the interactions can be derived
from others.  All possible interactions are defined by the power
set of $V$, that is, the set whose elements are all the possible
subsets of elements of $V$. If some interactions can be explained
in terms of others, then some groups of variables in $V$ are
independent from other groups, and the set that defines all
present interactions is smaller than the power set. To identify
the subsets of variables whose dependencies suffice to explain all
interactions, we propose different structured sets $\Omega =
\{U_1, U_2, \dots, U_\ell\}$, where each $U_i = \{X_{i_1}, \dots,
X_{i_k}\}$ is itself a set of variables that may or may not belong
to $V$. Each set $\Omega$ is a candidate explanation of the
statistical structure in $V$. Within the maximum entropy approach,
for each proposed $\Omega$ we calculate
\begin{equation}
\begin{array}{ll}
\Delta^{V}_{\Omega} &= D[p_{\Omega \cup V}:p_{\Omega}]\\\\
&=H_{\Omega}-H_{\Omega \cup V},
\end{array}
\label{eq:16}
\end{equation}
\noindent where we are using the notation described in the
previous section, so that $p_{\Omega}$ is the maximum entropy
distribution compatible with the marginals of the groups of
variables $U_1, U_2, \dots, U_\ell$ contained in $\Omega$, and
$p_{\Omega \cup V}$ is the maximum entropy distribution compatible
with the marginals of $U_1, \cdots, U_\ell, V$. If
$\Delta^{V}_{\Omega}$ is zero, then $p_{\Omega \cup
V}=p_{\Omega}$, and the joint probability of the variables $V$ can
be derived from $\Omega$. This means that the statistical
dependencies among the groups that compose $\Omega$ suffice to
explain the statistical structure among the groups that compose
$V$, even if the former contains interactions whose order is
smaller than the number of elements in $V$.

In the simplest example, we want to decide whether the statistical
structure in the pairwise marginal $p_{12} = p(X_1, X_2)$ may or
may not be explained by the univariate marginals $p_1 = p(X_1)$
and $p_2 = p(X_2)$. In this case, $V = \{X_1, X_2\}$ and $\Omega =
\{U_1, U_2\}$, with $U_1 = \{X_1\}, U_2 = \{X_2\}$.  When
calculating the union $\Omega \cup V$, we notice that here the
sign $\cup$ represents a union of marginals, not a union of sets.
The bivariate marginal $p_{12}$ contains the univariate marginals
$p_{1}$ and $p_{2}$, so $\Omega \cup V = V$. Hence,
\begin{equation}
\Delta^{12}_{1,2} = D[p_{12}:p_{1,2}] = I(X_1; X_2).
\label{eq:17bb}
\end{equation}
If $\Delta^{12}_{1,2} = 0$, the entire statistical structure
within $V$ is accounted for by the two independent variables $X_1$
and $X_2$.

In a more complex example, we may wish to determine whether the
statistical dependencies between the variables $X_1, X_2$ and
$X_3$ can be explained by just first and second order
interactions. We define $V = \{X_1, X_2, X_3\}$ and $\Omega =
\{U_1, U_2, U_3\}$, with $U_1 = \{X_1, X_2\}$, $U_2 = \{X_2,
X_3\}$, $U_3 = \{X_3, X_1\}$. The triple marginal $p_{123}$
contains all pairwise marginals $p_{12}, p_{23}$ and $p_{31}$, so
again, $\Omega \cup V = V$. Therefore,
\begin{equation}
\Delta^{123}_{12,13,23} = D[p_{123}:p_{12,13,23}]=D^{(3)}_{123}.
\label{eq:17b}
\end{equation}
\noindent If $\Delta^{123}_{12,13,23} = 0$, pairwise interactions
suffice to explain all the statistical structure in $V$.

A less ambitious goal would be to determine whether the
statistical dependence between $X_1$ and $X_2$ is mediated by a
third variable $X_3$. We hence define $V = \{X_1, X_2\}$, $\Omega
= \{U_1, U_2\}$, and $U_1 = \{X_1, X_3\}$, $U_2 = \{X_3, X_2\}$.
The union of marginals is now $\Omega \cup V = \{V, U_1, U_2\} \ne
V$, so in this case, $\Delta^{12}_{13,23}$ is given by
Eq.~(\ref{eq:17a}).

The set $\Omega$ constitutes a candidate explanatory model for the
statistical dependencies within $V$. The aim is to find the
simplest set $\Omega$ for which $\Delta^V_\Omega = 0$. The search
for such $\Omega$, however, has to be done within the power set of
the set that includes all the variables in the system, so the
number of candidate $\Omega$ sets grows exponentially with the
number of variables. Since for a large system the search becomes
computationally intractable, here we restrict the analysis to the
study of pairwise dependencies, that is, sets $V$ with just two
elements. Moreover, we search for explanatory models that attempt
to reproduce all the statistical structure in $V$ by means of
pairwise interactions with a third variable, as in
Eq.~(\ref{eq:17a}). A similar approach, but within a different
theoretical framework, has been proved useful in disambiguating
couplings in oscillatory systems \cite{kralemann14}. We define the
amount of irreducible interaction between the variables $X_i$ and
$X_j$ as the amount of statistical dependencies that remain
unexplained by the optimal minimal model, that is,
\begin{equation}
\begin{array}{ll}
\displaystyle \Delta^{ij} &\displaystyle = \min \left\{ \Delta^{ij}_{i,j} ,\min_k\{\Delta^{ij}_{ik,kj}\} \right\}\\\\
&\displaystyle =  \min \left\{ I_{ij} ,\min_k\{\Delta^{ij}_{ik,
kj}\} \right\}, \\\\ &\displaystyle =  \min \left\{ I_{ij}
,\min_k\{H_{ik,kj}-H_{ij, jk, ki}\}\right\}.
\end{array}
\label{eq:19}
\end{equation}
\noindent The index $k$ ranges through all the variables that do
not coincide with $i$ or $j$ ($k\neq i, k\neq j$). By defining
$\Delta^{ij}$ as a Kullback-Leiber divergence, its non-negativity
is ensured. Besides, the minimization in Eq.~(\ref{eq:19}) ensures
that $\Delta^{ij}$ is upper bounded by the mutual information,
that is, $\Delta^{ij}\leq I_{ij}$. Expanding
$\Delta^{ij}_{ik,kj}$,
\begin{equation}
\begin{array}{ll}
\Delta^{ij}_{ik,kj} &= H_{ik}+H_{kj}-H_k-H_{ij, jk, ki} \\\\
&= H_{ik}+H_{jk}-H_k-H_{ijk}+H_{ijk}-H_{ij, jk, ki} \\\\
&= I_{ij|k}-D^{(3)}_{ijk}.
\end{array}
\label{eq:21}
\end{equation}

Therefore, if there are not triple interactions within the whole
set of variables, then $\Delta^{ij}$ correspond to conditioning
the mutual information between $i$ and $j$ with every other
possible variable $k$, and looking for the minimum. We can rewrite
Eq.~(\ref{eq:19}) as
\begin{equation}
\begin{array}{ll}
\Delta^{ij} &\displaystyle = I_{ij}-\Theta\left(\max_k\left\{I_{ijk}+D^{(3)}_{ijk} \right\}\right) \\\\
&\displaystyle = I_{ij}-\Theta\left(\max_k\left\{I_{ij}+I_{jk}+I_{ki}-D^{(2)}_{ijk} \right\}\right) \\\\
\end{array}
\label{eq:22}
\end{equation}
\noindent where $\Theta(x)$ is the Heaviside step function. In
this sense, we are looking for a triplet that has maximal
redundancy, understanding redundancy as $\sum I-D^{(2)}$.

\section{Example of marginalization effects}
\label{app:example}

Consider four binary variables $X_i= \pm 1$, which can be thought
of as spins, with only pairwise interactions between $X_4$ and
each of the other three variables. The fourth variable is in the
up state with probability $(1 + {\rm e}^{- 2 \beta})^{-1}$. Here
we focus in negative $\beta$ values, which favor the down state.
The joint probability can be written as a log-linear model
\cite{agresti,amari01}
\begin{equation}
\begin{array}{ll}
\displaystyle \log p(x_1, x_2, x_3, x_4) &= \beta x_4 + x_1 x_4 + x_2 x_4 + x_3 x_4 - \psi \\\\
 &= (\beta + x_1 + x_2 + x_3) x_4 - \psi
\end{array}
\label{eq:24}
\end{equation}
\noindent where $\beta < 0$ is the field acting on $X_4$, and
$\psi$ is the normalization constant. Marginalizing over $X_4$, we
obtain
\begin{equation}
\begin{array}{ll}
\displaystyle p(x_1, x_2, x_3) &\displaystyle = \frac{\cosh(\beta
 + x_1 + x_2 + x_3)}{\sum_{\textbf{x}'} \cosh(\beta + x'_1 + x'_2 + x'_3)}.
\end{array}
\label{eq:25}
\end{equation}
With this probability we are able to calculate the interactions
$\Delta_{123}$, $D^{(2)}_{123}$ and $D^{(3)}_{123}$ as a function
of $\beta$.
\begin{figure}[ht]
    \begin{center}
    \includegraphics[width=8cm, clip=true]{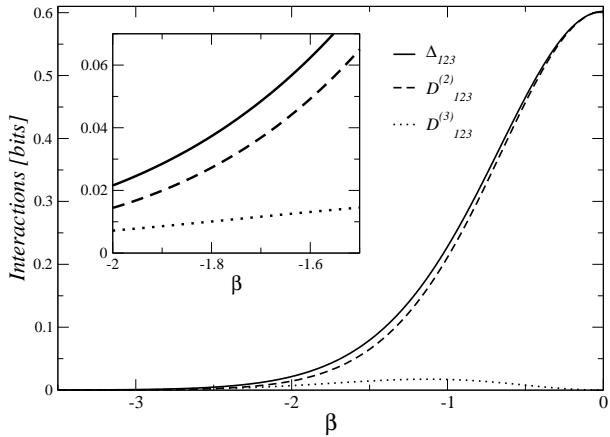}
    \end{center}
\caption{\footnotesize{Interactions $\Delta_{123}$,
$D^{(2)}_{123}$ and $D^{(3)}_{123}$ as a function of the field
$\beta$ acting on $X_4$.}} \label{f:mar}
\end{figure}

In Fig.~\ref{f:mar} we see the multi-information $\Delta_{123}$,
the amount of pairwise interactions in the triplet
$D^{(2)}_{123}$, and the triple information $D^{(3)}_{123}$ as a
function of the field $\beta$ acting on $X_4$. As stated above,
$\Delta_{123} = D^{(2)}_{123} + D^{(3)}_{123}$. All of these
quantities are obtained from the marginal probabilities $p(x_1,
x_2, x_3)$ given by Eq.~(\ref{eq:25}) (see Appendix
\ref{app:solp2}). When the field is strong ($\beta \rightarrow
-\infty$) the total amount of interaction vanishes, as all spins
align in the down state. For small values of the field, the amount
of interactions is large, and can be explained almost entirely by
pairwise dependencies. However for intermediate values of the
field (see inset of Figure~\ref{f:mar}), which corresponds to the
fourth spin aligned downwards most of the time, the triple
information is crucial to understand the structure of dependencies
within the group of remaining variables. In this paper we argue
that in the case of written language, the topics or latent
variables that affect the occurrence of words are likely to
present the same kind of behavior, that is, they tend to be
inactive most of the time. And when they are active, they tend to
favor the occurrence of specific groups of words.

\section{\label{app:sign}Significance test}

We want to assess whether a probability
distribution of three variables $p(\textbf{x})$ is explained or
not by the simpler maximum entropy model $p^{(2)}(\textbf{x})$,
obtained after measuring only the pairwise marginal probabilities.
That is, taking the maximum entropy model as the null hypothesis
$H_0$, and considering as the alternative hypothesis $H_1$ the one
in which there is a triple dependency, we want to calculate the
plausibility of the distribution $p(\textbf{x})$. In statistics a
usual way of comparing two models, one of which is nested within the
other, is a likelihood ratio test.

If we take $N$ samples, then the likelihood ratio $\lambda$ is given by
\begin{equation}
\begin{array}{ll}
\lambda &\displaystyle = \frac{P(\textbf{x}_1,...,\textbf{x}_N|H_1)}{P(\textbf{x}_1,...,\textbf{x}_N|H_0)} \\\\
&\displaystyle = \frac{\prod_{i=1}^N p(\textbf{x}_i)}{\prod_{i=1}^N p^{(2)}(\textbf{x}_i)}.
\end{array}
\label{ea1:00}
\end{equation}
Considering $N\rightarrow \infty$ and using Sanov's theorem \cite{cover}, it follows
\begin{equation}
\log(\lambda)=N D[p:p^{(2)}].
\label{ea1:01}
\end{equation}
In addition, the result by Wilks \cite{wilks38} implies that, neglecting terms of order $N^{-1/2}$,
\begin{equation}
2\log(\lambda)= \chi^2_d,
\label{ea1:02}
\end{equation}
\noindent that is, the logarithm of the likelihood tends to a 
chi-square distribution, where the number of degrees of freedom $d$ 
equals the difference in the numbers of parameters between the models.
Combining these two results, we conclude that under the null hypothesis,
\begin{equation}
D[p:p^{(2)}]= \frac{\chi^2_1}{2 N},
\label{ea1:03}
\end{equation}
\noindent where the chi-square distribution has one degree of freedom. Taking a significance of $\alpha=0.1\%$ and $N=512$,
we reject the null hypothesis if $D[p:p^{(2)}]\gtrsim 0.01$ bits.

An analogous analysis is done when evaluating the significance of $D[p_{ij,ik,jk}:p_{ik,jk}]$, with the same result.

\section{\label{app:err}Error estimation}
The estimation of the error of our measures is done by a bayesian
approach \cite{samengo02}. Estimation problems are dominated by
finite sampling in the probabilities of the different states.

On the one side, we have the true probability $\textbf{q}$
governing the outcome of the experiment, whose coordinates refers
to the $S$ possible states of the system (in our case to the eight
states for three binary variables). On the other side, there is
the frequency count $\textbf{f} = n_i / n$, where $n_i$ is the
number of times the state $i$ occurs, and $N$ is the total number
of measurements. The probability of measuring $\textbf{f}$ given
that the data are governed by $\textbf{q}$ is the multinomial
probability
\begin{equation}
\displaystyle p(\textbf{f}|\textbf{q})=N! \prod_i \frac{q_i^{n_i}}{n_i!}=N! \prod_i \frac{q_i^{N f_i}}{(N f_i)!}.
\label{ea2:00}
\end{equation}

We have no access to $\textbf{q}$, we can only measure
$\textbf{f}$. We therefore need the probability that the true
distribution be $\textbf{q}$ given that $\textbf{f}$ was measured,
that is, the probability density $P(\textbf{q}|\textbf{f})$.
Through Bayes' rule,
\begin{equation}
\begin{array}{ll}
\displaystyle P(\textbf{q}|\textbf{f})&\displaystyle =\frac{p(\textbf{f}|\textbf{q})P(\textbf{q})}{p(\textbf{f})}\\\\
&\displaystyle =\frac{\exp\left(-N D[\textbf{f}:\textbf{q}]\right)P(\textbf{q})}{\textit{Z}}
\end{array}
\label{ea2:01}
\end{equation}
\noindent where $P(\textbf{q})$ is the prior probability
distribution for $\textbf{q}$, and $\textit{Z}$ is the
normalization over the domain of $\textbf{q}$. For the estimation
of the error, and in the limit of a large number of samples, the
result does not depend on the choice of the prior, as we show
below.


If we need to estimate some function of the probabilities
$W(\textbf{q})$, the variance of the estimate is
\begin{equation}
\displaystyle \sigma^2_W = \langle W^2 \rangle - \langle W \rangle^2,
\label{ea2:03}
\end{equation}
\noindent where the average is over $P(\textbf{q}|\textbf{f})$. In
our case, we are interested in the triple information
$W(\textbf{q})=D[\textbf{q}:\textbf{q}^{(2)}]$, where
$\textbf{q}^{(2)}$ is the maximum entropy probability compatible
with the second-order marginals.

From \cite{samengo02} it follows that, in the limit $N \gg S$ and to a first order in $1/N$,
\begin{equation}
\begin{array}{lll}
\sigma^2_W & \approx &\displaystyle \sum_i \left. \left(\frac{\partial W}{\partial q_i} \right)^2\right|_f \frac{f_i(1-f_i)}{N} \\\\
&&\displaystyle -2 \sum_i \sum_{j<i} \left. \left(\frac{\partial W}{\partial q_i} \frac{\partial W}{\partial q_j} \right)\right|_f \frac{f_i f_j}{N} +O(N^{-2})\\\\
&=&\displaystyle \nabla_{q}W^t \cdot \varSigma \cdot \nabla_{q} W,
\end{array}
\label{ea2:04}
\end{equation}
\noindent where the covariance matrix of the probabilities $\varSigma$ is
\begin{equation}
\varSigma_{ij}= \left\{
\begin{array}{ll}
\displaystyle \frac{f_i(1-f_i)}{N} &\text{ if }i=j \\\\
\displaystyle -\frac{f_i f_j}{N} &\text{ if }i\neq j
\end{array}
\right.
\label{ea2:05}
\end{equation}
Due to finite sampling, the frequencies $f_i$ may fluctuate. From
Eq.~(\ref{ea2:04}) we see that we only need the covariance matrix
and the gradient of $W(\textbf{q})$ evaluated in $\textbf{f}$ in
order to transform the variance of the vector $\textbf{f}$ along
different directions of the simplex into variance in $W$. It is
important to notice that the error in $W$ is of order
$1/\sqrt{N}$, which means that if we want to reduce the error by
half, we need to increase the number of samples fourfold.

In our case the gradient $\nabla_{q} W$ is difficult to calculate,
but we can obtain the result from Eq.~(\ref{ea2:04}) numerically.
Given the frequency $\textbf{f}$, first we calculate the
eigenvalues and eigenvectors from the covariance matrix
$\varSigma$ given by Eq.~(\ref{ea2:05}). One non-degenerate
eigenvector is orthogonal to the simplex, and has a zero
eigenvalue. The remaining eigenvectors $\textbf{v}_k$ belong to
the simplex and all have positive eigenvalues $\sigma^2_k$, equal
to the variances in the corresponding directions. Finally, making
a small change $\epsilon$ in the frequencies along these
directions, we obtain the change $\Delta W_k=W(\textbf{f}+\epsilon
\textbf{v}_k)-W(\textbf{f})$, so that
\begin{equation}
\displaystyle \sigma^2_W = (\Delta W)^2 \approx
\frac{1}{\epsilon^2} \sum_{k=1}^{S-1} \Delta W_k^2 \sigma_k^2,
\label{ea2:06}
\end{equation}
\noindent where every $\sigma_k^2$ is in the order of $1/N$.

\begin{figure}[ht]
    \begin{center}
    \includegraphics[width=8cm, clip=true]{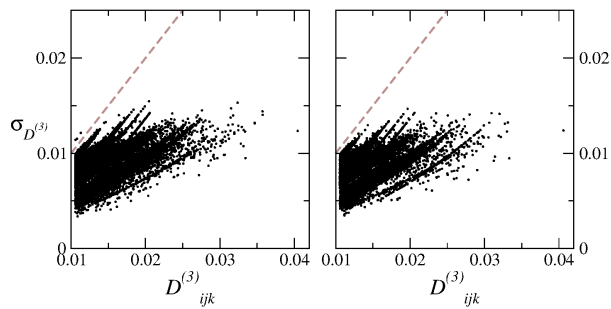}
    \end{center}
\caption{\footnotesize{Standard deviation of the triple
information $D^{3}_{ijk}$ as a function of the $D^{3}_{ijk}$, for
the triplets that satisfy $D^{3}_{ijk}>0.01$. A: OS. B:
AM. The dashed line indicates the identity.}} \label{f:error}
\end{figure}

Figure \ref{f:error} shows the standard deviation of $D^{3}_{ijk}$
obtained by this method as a function of $D^{3}_{ijk}$ for the
triplets that satisfy $D^{3}_{ijk}>0.01$, for both books. The
error lies between $0.005$ bits and $0.01$ bits.

\bibliography{paper_ddi_5}

\end{document}